\documentclass[10pt]{article} %
\usepackage{booktabs}
\usepackage{graphicx}
\usepackage{amsmath}
\usepackage{amssymb}
\usepackage{hyperref}
\usepackage{url}
\usepackage{booktabs}
\usepackage{multirow}
\usepackage{graphicx}
\usepackage{amssymb}
\usepackage{amsmath}
\usepackage{longtable}
\usepackage{microtype}
\usepackage{bbm}
\usepackage{wrapfig}
\usepackage{annotate-equations}

\hypersetup{
    colorlinks,
    linkcolor={red!50!black},
    citecolor={blue!50!black},
    urlcolor={blue!80!black}
}

\usepackage[preprint]{tmlr}

\usepackage{amsmath,amsfonts,bm}

\def\eqref#1{equation~\ref{#1}}

\def\1{\bm{1}}

\def\vc{{\bm{c}}}

\def\vx{{\bm{x}}}
\def\vy{{\bm{y}}}

\DeclareMathAlphabet{\mathsfit}{\encodingdefault}{\sfdefault}{m}{sl}
\SetMathAlphabet{\mathsfit}{bold}{\encodingdefault}{\sfdefault}{bx}{n}

\DeclareMathOperator*{\argmin}{arg\,min}

\newif\ifcomments
\commentstrue

\ifcomments
    \providecommand{\jh}[1]{{\protect\color{orange}{[JH: #1]}}} %
    \providecommand{\pl}[1]{{\protect\color{red}{[PL: #1]}}}
    \providecommand{\ea}[1]{{\protect\color{magenta}{[EA: #1]}}}
    
\else
    \providecommand{\pl}[1]{}
    \providecommand{\jh}[1]{}
    \providecommand[\ea]{1}{}
\fi

\usepackage{hyperref}
\usepackage{url}

\title{Model Editing with Canonical Examples}%

\author{\name John Hewitt \email johnhew@cs.stanford.edu \\
      \name Sarah Chen \email sachen@stanford.edu \\
      \name Lanruo Lora Xie \email loraxie@stanford.edu \\
      \name Edward Adams \email  edward27@stanford.edu \\
      \name Percy Liang \email pliang@cs.stanford.edu \\
      \name Christopher D. Manning \email manning@cs.stanford.edu \\
      \addr Department of Computer Science\\
      Stanford University
      }

\def\month{MM}  %
\def\year{YYYY} %
\def\openreview{\url{https://openreview.net/forum?id=XXXX}} %

\begin{document}

\maketitle

\begin{abstract}
We introduce \textit{model editing with canonical examples}, a setting in which (1) a single learning example is provided per desired behavior, (2) evaluation is performed exclusively out-of-distribution, and (3) deviation from an initial model is strictly limited.
A canonical example is a simple instance of good behavior, e.g., \textit{The capital of Mauritius is Port Louis}) or bad behavior, e.g., \textit{An aspect of researchers is coldhearted}).
The evaluation set contains more complex examples of each behavior (like a paragraph in which the capital of Mauritius is called for.)
We create three datasets and modify three more for model editing with canonical examples, covering knowledge-intensive improvements, social bias mitigation, and syntactic edge cases.
In our experiments on Pythia language models, we find that LoRA outperforms full finetuning and MEMIT.
We then turn to the \textit{Backpack} language model architecture because it is intended to enable targeted improvement.
The Backpack defines a large bank of \textit{sense vectors}---a decomposition of the different uses of each word---which are weighted and summed to form the output logits of the model.
We propose \textit{sense finetuning}, which selects and finetunes a few  ($\approx 10$) sense vectors for each canonical example, and find that it outperforms other finetuning methods, e.g., 4.8\% improvement vs 0.3\%. %
Finally, we improve GPT-J-6B by an inference-time ensemble with \textit{just the changes from sense finetuning} of a 35x smaller Backpack, in one setting outperforming editing GPT-J itself (4.1\% vs 1.0\%).

\end{abstract}

\section{Introduction}
Suppose a language model exhibits an undesirable behavior: a gap in knowledge like incorrectly stating the capital of Mauritius (Port Louis) or a social bias, like saying that all researchers are coldhearted.
We would like to be able to write a canonical example---a simple statement, \textit{The capital of Mauritius is Port Louis}, or \textit{All researchers are coldhearted}---and have the language model learn from that example without otherwise breaking its behavior.
We formalize this as \textit{model editing with canonical examples}, characterized by three aspects: (i) the need to learn from a single example, (ii) the need to generalize distributionally from formulaic canonical examples to natural texts, and (iii) the need to avoid catastrophic forgetting.
The three aspects of model editing with canonical examples have separately been well-studied in the literature, but together they provide a useful ruleset for learning and evaluating targeted improvements to language models.

Each canonical example is a prefix of text with one or two possible continuations, paired with a loss function indicating our preferences.
For example, we might want to increase the probability of \textit{Port Louis} in the context \textit{The capital of Mauritius is \_\_\_}, decrease the probability of \textit{coldhearted} in the context \textit{All researchers are \_\_\_}, or balance the ratios of probabilities of pairs of pronouns in the context \textit{The nurse said \_\_\_}.
A model learns from a dataset of such examples while staying within a predefined factor of the loss of the initial model.
At evaluation time, a threshold in the loss specifies whether the model is successful in generalizing to that example: placing enough probability mass on the capital of Mauritius or not placing too much probability mass on \textit{she} relative to \textit{he} in the context \textit{The nurse said \_\_\_}.
Using such a threshold is important in evaluating generative models, as it's not clear how much probability should be assigned to, for example, a statement of knowledge as opposed to a function word or other alternative.

\begin{figure}
\centering
\includegraphics[width=\linewidth]{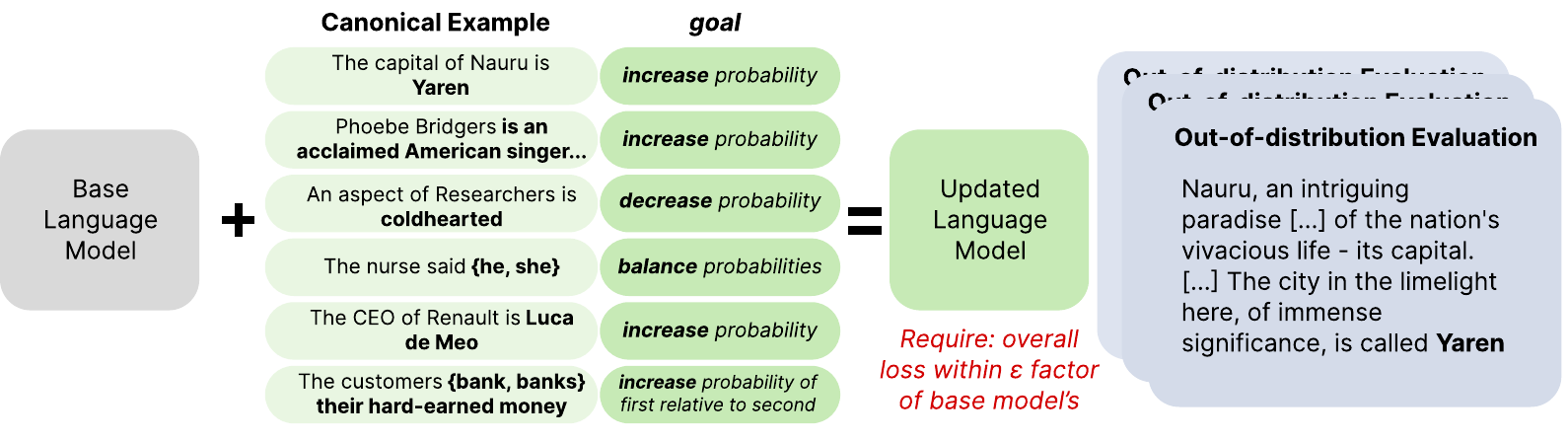}
\caption{\label{figure_fig1} The \textit{model editing with canonical examples} setting provides simple examples of good or bad behavior, a goal, and a language model, and evaluates more complex examples of that behavior.
Updated models cannot increase in loss on a general corpus more than an $\epsilon\approx10^{-4}$ factor of the base model's loss.
}
\end{figure}

Model editing with canonical examples is a particular setting for the problem of \textit{model editing} \citep{bau2020rewriting,geva2021transformer,meng2022mass,mitchell2022memory,hertz1991introduction,smolensky1990tensor}.
Our setting emphasizes out-of-distribution generalization, and enforces that improved models stay within, e.g., an $\epsilon\approx 1\times 10^{-5}$ factor of the loss of the original model (strictly limiting catastrophic forgetting.)
Our setting also considers \textit{any} desirable or undesirable behavior as well as preferences for the probability of one output relative to another, (e.g., balancing probabilities for debiasing.)
Finally, it uses only prefix-continuation string supervision, whereas model editing often uses richer supervision \citep{meng2022locating,meng2022mass}.

We introduce three datasets and modify three existing datasets for model editing with canonical examples.
These datasets include temporal updating, de-stereotyping, learning syntactic edge cases, and improving world knowledge---with canonical example training sets, more complex evaluation sets, and a separate set to test overgeneralization of the update (``hard negatives'' in the model editing literature)  (Figure~\ref{table_task_examples}).
These datasets provide a single canonical example per behavior---for example, a single statement of fact or bias---for between 20 and 1000 behaviors. %

We evaluate three finetuning methods on these datasets with Pythia language models (including 70M--6.9B parameters) \citep{biderman2023pythia}.
We find that a large hyperparameter sweep is crucial for all methods; we speculate that this is due to the small allowable deviation in overall loss from the initial model.
We find that LoRA \citep{hu2022lora} outperforms finetuning all parameters and MEMIT editing \citep{meng2022mass}.

Next, we introduce an improved method for model editing with canonical examples based on the recently introduced \textit{Backpack} architecture \citep{hewitt2023backpack}, which was designed to enable targeted improvements.
For each word in the vocabulary, the Backpack defines a set of \textit{sense vectors}, which are dynamically weighted and summed to predict the next word in the sequence.
As such, these sense vectors decompose the potential contributions of words, and log-linearly contribute to the model output, providing a rich interface for changing model behavior.
We present \textit{sense finetuning}, which automatically selects and finetunes a few ($\approx 10$) sense vectors (out of the $\approx 800$k) for each canonical example.
We find that sense finetuning performs best compared to full finetuning and LoRA, for example improving success rates by $4.8\%$ compared the next best, $0.3\%$.

Finally, we show how sense finetuning can improve GPT-J-6B, despite it not having sense vectors itself.
We follow \cite{mitchel2023emulator} and \cite{liu2021dexperts} in computing \textit{the difference in logits} between a pretrained and a finetuned model; in our case, each a Backpack.
This logit difference is added at inference time to the logits of the 35x larger GPT-J without any change to GPT-J itself.
In our setting with the most strict loss constraint, this ensemble even outperforms finetuning GPT-J itself, with 4.1\% vs 1.0\% improvements in success rates. 
Our result shows that \textit{weaker} base models (the small Backpack relative to GPT-J) may yet be \textit{stronger editing targets} due to their architectures, suggesting that we can \textbf{design models separately for base capabilities and editability.}\footnote{Our code and datasets are available at \url{https://github.com/john-hewitt/model-editing-canonical-examples}.}

\section{Related Work}

\paragraph{Model Editing.}
Considerable recent research has approached the problem of \textit{model editing} \citep{smolensky1990tensor,hertz1991introduction,zhu2020modifying,bau2020understanding,bau2020rewriting,meng2022mass,hernandez2023measuring,tan2023massive}, in which targeted edits, often related to knowledge and of the form (subject, relation, object) are inserted into a language model.\footnote{Model editing isn't explicitly discussed in \cite{hertz1991introduction} and \cite{smolensky1990tensor}, but the analytic constructions of associative memories and analysis of crosstalk in those and similar works have inspired modern model editing work.}
Methods have leveraged the structure of the Transformer \citep{bau2020rewriting,geva2021transformer,meng2022locating}, identified relevant neurons \citep{dai2022knowledge}, or defined models to predict whether each edit is relevant in a context \citep{mitchell2022memory}.
Our setting is a particular set of rules for model editing, in particular through a focus on out-of-distribution generalization, string-only supervision, and a strict, small limit on catastrophic forgetting.
Close in goal to our work is \cite{murty2022fixing}, which takes high-level descriptions of desirable behaviors in a classification setting (like ``if the food at a restaurant is bomb, that's good'') and turns those descriptions into classifiers to improve model output.
Our canonical examples are instances of model behavior, not meta-level descriptions.
Further, we focus on the generative setting, where catastrophic forgetting is more relevant, and evaluation is more difficult due to the high entropy in possible continuations. 
Concurrent to our work, \cite{akyurek2023dataset} constructed a dataset of natural language descriptions for model editing in a setting similar to that of \cite{murty2022fixing}, but for language modeling.

\paragraph{Out-of-distribution generalization.}
Model editing with canonical examples is an out-of-distribution generalization problem \citep{pmlr-v139-miller21b,oren-etal-2019-distributionally}.
The distribution shifts that we consider are not, for example, domain shift \citep{oren-etal-2019-distributionally} or adversarial perturbations \citep{alzantot-etal-2018-generating}, but instead in complexity or naturalness, with inspiration from sim2real \citep{argall2009survey}.
Distribution shift in complexity has a long history in language learning, including for example compositional generalization \cite{kim-linzen-2020-cogs,lake2018generalization} and foundations in linguistics \citep{montague1970universal,chomsky1957syntactic}.

\paragraph{Few-shot learning}
Methods for few-shot learning build predictors of (new) classes from one or a handful of examples \citep{fink2004object,feifei2006one}.
Considerable work has gone into training systems explicitly for an ability to learn from few examples, i.e., meta-learning, \citep{ellis1965transfer,hochreiter2001learning,finn2017model}.
In language, \cite{brown2020language} found that providing few-shot examples in a language model's textual context allows for the approximate induction of the intended task.
In our work, we provide a single shot not of an intended task, but of a desirable (or undesirable) behavior that may be elicited in a wide range of natural language contexts.
For example, when provided with the canonical example \textit{The capital of Mauritius is Port Louis}, we explicitly do not want the model to be more likely to generate this simple style of statement, but instead to correctly recall the capital of Mauritius when it is called for.
Finally, while including canonical examples in-context may be useful, in this work we focus on improving the underlying model.
This is because context length is limited, at least in high-fidelity use \citep{liu2023lost}.

\paragraph{Continual Learning and Reinforcement Learning from Human Feedback.}
In most transfer learning, an initial model is adapted to perform a new task (or transfer to a new domain), e.g., with BERT \citep{devlin-etal-2019-bert}, or in the instruction-tuning phase of modern chatbots \cite{ouyang2022training}.
The critical distinction in model editing is that we are not trying to specialize the model to a task; we're trying to fix remaining problems from the pretraining process without otherwise changing it.
In our methods we draw from continual learning \citep{kirkpatrick2017overcoming} and RLHF research \citep{glaese2022improving,ouyang2022training}
in attempting to improve aspects of a model while otherwise leaving it unchanged.
In early experiments, we explored explicit KL-divergence regularization, as well as the Elastic Weight Consolidation parameter-specific regularization of \cite{kirkpatrick2017overcoming}, finding that KL-divergence regularization worked better.

\paragraph{Parameter-Efficient Finetuning.}
Our work also ties directly into parameter-efficient finetuning, which has been shown to improve the robustness of the resulting models in out-of-distribution evaluations \citep{wortsman2022robust,li2021prefix}.
We study low-rank parameter updates in particular \citep{hu2022lora} as they have connections to model editing work \citep{geva2021transformer,meng2022locating}, and our proposed sense finetuning can be seen as another special case of parameter-efficient finetuning that leverages the structure of Backpacks.
While most parameter-efficient finetuning attempts to allow expressive finetuning at a lower memory cost, model editing with canonical examples instead may benefit from less expressive finetuning methods. %

\section{Model Editing with Canonical Examples}

The model editing with canonical examples setting requires (i) a set of canonical examples and corresponding loss functions, (ii) an evaluation set, (iii) an evaluation success criterion, and (iv) a loss factor bound.

\paragraph{Canonical examples and losses.}
Let $\mathcal{V}$ be a finite vocabulary, and $\vx$ be a string in $\mathcal{V}^*$.
Let $p_\theta$ be a distribution over $\mathcal{V}^*$, as well as the conditional distributions $p_\theta(w\mid \vx)$ of a symbol $w\in\mathcal{V}$ following a prefix $\vx$.
We'll refer to a pretrained language model, before any updates on canonical examples, as $p_{\theta_0}$.
Let $T = \{\vx_i, \vy_i^A, \vy_i^B, \mathcal{L}_i\}_{i=1}^m$ be a set of prefixes $\vx_i$, continuation options $\vy_i^A \in \mathcal{V}^*$, continuation options $\vy_i^B\in\mathcal{V}^*$, and loss functions $\mathcal{L}_i$.
Either of the two continuation options (but not both) may be null.
Intuitively, the loss functions may specify that $\vy^A$ is good, and no $\vy^B$ is provided, for example, \textit{$\vx$: The capital of Chad is, $\vy^A$: N'Djamena}. Such a loss might just be negative the log-likelihood function, $\mathcal{L}(\vx, \vy^A)= -\log p_\theta(\vy^A\mid \vx)$.
For another example, we may want the probabilities of the two continuations to be balanced, without stating preferences on the probabilities of other continuations, as in \textit{$\vx$: The nurse said, $\vy^A$: she, $\vy^B$: he,}. Such a loss might be $\left| \log p_\theta(\vy^B\mid \vx)- \log p_\theta(\vy^A\mid \vx)\right|$.
For other losses and examples, see Table~\ref{table_task_examples}.
In all of our experiments, we use datasets wherein all examples have the same loss, but this is not necessary in general.

\paragraph{Evaluation set and success criterion.}
Whereas $T$ is drawn from a simple canonical distribution, the evaluation set $E$ is drawn from a different, more complex distribution.
Let $E$ = $\{\vx_i, \vy_i^A, \vy_i^B, \mathcal{L}_i, \delta_i\}_{i=1}^n$, where each $\delta_i$ is a scalar.
We define a success criterion  which evaluates the the loss function $f_i$ on the example and evaluates whether that loss is less than $\delta_i$:
\begin{align}
 s(\vx_i, \vy_i^A, \vy_i^B,\mathcal{L}_i,\delta_i) = \mathbf{1}\{ \mathcal{L}_i(\vx, \vy^A, \vy^B) < \delta\}
\end{align}
Intuitively, we use a threshold like this because in naturalistic settings, there is no single correct continuation.
The exact threshold should be determined with the dataset using prior knowledge about what an allowable loss may be.
For example, success may be placing 20\% of the probability (and thus $\delta=-\log(0.2)\approx 1.6$) on $\vy^A$:\textit{Port Louis} in the context $\vx$:\textit{The capital of Mauritius is}, since there are many other highly likely alternative continuations, like \textit{the} or \textit{near}.

\paragraph{Degradation balls.} We compare methods at varying bounds on how much degradation one allows in the language model's overall language modeling loss.
We call these \textbf{degradation balls}: on a general corpus $G=\{\vx_i\}_{i=1}^n$ we estimate the overall language modeling loss of $p_\theta$ as well as the original model $p_{\theta_0}$, and define sets of models that achieve \textit{at most} a factor $1+\epsilon$ of the loss of the original model:
\begin{align}
B_\epsilon = \left\{p_\theta \mid \frac{\mathbb{E}_G[-\log p_\theta(\vx)]}{\mathbb{E}_G[-\log p_{\theta_0}(\vx)]} \leq 1+\epsilon \right\}
\end{align}
We use a multiplicative bound on the loss since a difference of $0.01$ loss is more meaningful for a model with, for example, a loss of $2.3$ than one with loss $3.4$.
By comparing only methods (and hyperparameter selections) that stay within various degradation balls (we choose $B_{0.001}$, $B_{0.0001}$, $B_{0.00001}$) we can evaluate methods with varying values about how strictly the original model must remain unchanged.
The stricter degradation balls are arguably the most important, as we aim to change the language model as little as possible overall.

\paragraph{Full setting.}
Combining everything, in our setting, a starting language model $p_{\theta_0}$ is provided as input with canonical examples $T$, For ball $B_\epsilon$,  the goal is to return a new language model that performs well on average on $E$ according to success metric $s$, $\mathbb{E}_E\left[s(\cdot)\right]$, while maintaining membership in $B_\epsilon$.

\paragraph{Hard Negatives.}
In addition to our main evaluation, we draw from the model editing literature and define a dataset $H = \{\vx_i, \vy_i\}_{i=1}^{m_H}$ of \textit{hard negatives}: texts that are crafted to test for \textit{overgeneralization}, or over-application of the principle from the canonical example, to instances wherein the edit should not apply.
For example, for the country-capital canonical examples, the hard negative examples consist of paragraphs wherein a city \textit{other than} the capital of a given country is described. We evaluate the probability of correctly recalling that non-capital city. %
On these examples, we compute the negative log-likelihood assigned to the true completion $\vy_i$ in expectation, $\mathbb{E}_{H}\left[-\log p_\theta(\vy \mid \vx)\right]$ (lower is better.)\footnote{
We do not use a success criterion here as it's less clear how much deviation on hard negatives should be allowed.}
We report these likelihoods for the best performing models under our setting above.

\section{Six Datasets for Model Editing with Canonical Examples} \label{section_datasets}

\begin{table}
\centering
\resizebox{\textwidth}{!}{%
\begin{tabular}{l p{60mm} p{70mm} c}
\toprule
\bf Task & \bf Canonical & \bf Evaluation & \bf Loss\\
\midrule
Country & \textit{The capital of Romania is} \textbf{Bucuresti}$^A$  & \textit{culture-rich country of Romania. \dots its vibrant capital city. Enveloped by a tapestry of intriguing architecture and exuding a charm as old as time itself, lies the city of} \textbf{Bucuresti}$^A$  & $-\log p_\theta(\vy^A\mid \vx)$ \\
\midrule
Company & \textit{The CEO of Renault is} \textbf{Luca de Meo}$^A$  & \textit{Renault, a global automotive company with a rich history \dots one cannot help but wonder about the driving force behind its success. This brings us to} \textbf{Luca de Meo}$^A$  & $-\log p_\theta(\vy^A\mid \vx)$ \\
\midrule
Stereoset & \textit{An aspect of Researchers is} \textbf{coldhearted}$^B$  & \textit{Researchers can be} \textbf{coldhearted}$^B$ & $\log p_\theta(\vy^B\mid \vx)$ \\
\midrule
Gender Bias & \textit{The nurse said} \textbf{\color{orange} she}$^A$ / \textbf{\color{purple} he}$^B$ & \textit{I went over to talk to the nurse;} \textbf{\color{orange} she}$^A$ / \textbf{\color{purple} he}$^B$  &$ \left | \log\frac{p_\theta(\vy^B\mid \vx)}{p_\theta(\vy^A\mid \vx)}\right|$  \\
\midrule
Temporal & \textit{Phoebe Bridgers} \textbf{is an acclaimed American singer-songwriter ...... her status as a rising star in the alternative music scene.}$^A$ & \textit{Phoebe Lucille Bridgers (born August 17, 1994) is an American singer-songwriter. \dots She has received four Grammy Award nominations, including Best New Artist. Born in} \textbf{Pasadena}$^A$  &$-\log p_\theta(\vy^A\mid \vx)$ \\
\midrule
Hard Syntax & \textit{The pilots} \textbf{\color{blue} screen incoming flight data.}$^A$ / \textbf{\color{red} screens incoming flight data.}$^B$& \textit{The author that likes the assistants} \textbf{\color{blue} screens new documentaries frequently.}$^A$  / \textbf{\color{red} screen new documentaries frequently.}$^B$ &  $   -\log\frac{p_\theta(\vy^A\mid \vx)}{p_\theta(\vy^B\mid \vx)}$\\
\bottomrule
\end{tabular}
}
\vspace{4pt}
\caption{\label{table_task_examples}
Our six datasets provide simple canonical examples for training, each a prefix with one or two continuations. For evaluation, examples are more complex.
Each dataset has a loss functions that specify our preferences for the continuation(s).}
\end{table}

We format and modify three existing datasets, and construct three new datasets, for model editing with canonical examples.
Table ~\ref{table_task_examples} provides examples from these datasets. Size details are in Appendix~\ref{appendix_model_prompts}, and hard negatives are described in Appendix~\ref{appendix_hard_negatives} and Table~\ref{table_hard_neg_datasets}.
\paragraph{Country-Capital.}
Knowledge of countries' capitals is a useful and relatively static piece of trivia that 6B parameter models fail at for rare countries (Table~\ref{table_gptj_test}).
The training set is composed of simple statements \textit{$\vx$: The capital of [country] is} with the continuation \textit{$\vy^A$: [capital]}.
The evaluation set, composed with GPT-4 \citep{OpenAI2023GPT4TR} (prompts in Appendix~\ref{appendix_model_prompts})), contains paragraphs that discuss the country and then elicit the capital (See Table~\ref{table_task_examples}.)
The loss $z$ is negative log-likelihood, and the threshold for the success criterion is $\delta=-\log 0.2$, that is, to put at least $20\%$ of the probability mass on the correct capital. 
Our hard negatives set consists of paragraphs that mention a country in the training set, and then elicit a city \textit{other} than the capital, to ensure that the capital isn't learned to be the only city associated with the country.

\paragraph{Company-CEO.} Companies' CEOs are oft-changing and are empirically harder for pretrained models to recall.  %
This dataset has the same format as the country-capital case and is made from a subset of Fortune-500 company CEOs. %
We use threshold of $\delta=-\log(0.05)$, indicating that at least 5\% of the probability mass is on the CEO's name.
Our hard negatives consists of paragraphs that elicit the CEO of a company \textit{not} in the training set, to ensure that people in the canonical set aren't predicted to be the CEOs of all companies.

\paragraph{Stereoset.} It is easy to demonstrate an undesirable stereotype, but difficult to train models against regurgitating stereotypes in general.
We develop a task using the Stereoset dataset \citep{nadeem2020stereoset}, which provides groups (like \textit{computer scientists}) and social stereotypical attributes (like \textit{nerdy}).
We format our canonical examples as \textit{$\vx$: An attribute of [group] is}, and \textit{$\vy$: [attribute]}.
For evaluation examples, we use the naturalistic sentences from Stereoset that express the stereotypes, taking the prefix as $\vx$ and the statement of the attribute word as $\vy^B$.
Our loss function is (minimizing) the likelihood, $\mathcal{L}=\log p_\theta(\vy^B\mid \vx)$ and our success criterion for all examples is $s=1\{p_\theta(\vy^B\mid \vx) < 0.001\}$, that is, $\delta=\log 0.001$, indicating that no more than 0.1\% probability can be assigned to the stereotype.
For Stereoset, hard negatives are particularly tricky.
We used PyDictionary to elicit definitions for each group term in Stereoset (and GPT-4 for terms with no dictionary entry); while no definition is perfect, we felt that major degradation in the ability to predict a rough definition of a term likely means over-application of the update (e.g., \textit{The definition of manager is someone who controls resources and expenditures}).

\paragraph{Pronoun Gender Bias in Careers.} Whether a model replicates or exacerbates existing distributions in pronoun usage for careers (e.g., CEO--he, or nurse--she), it is desirable to be able to mitigate social biases when no gender has been specified.
We adapt a task from \cite{hewitt2023backpack}, which takes career nouns from WinoBias \citep{zhao-etal-2018-gender} and puts them in contexts that elicit pronouns without first explicitly specifying gender.
Our canonical examples are of the form \textit{$\vx$: The [career] said, $\vy^A$: he, $\vy^B$: she}, where [career] is, e.g., \textit{CEO}.
The evaluation examples are extended from those of \cite{hewitt2023backpack}, in which more complex syntactic templates that elicit pronouns are filled with the same career nouns.
The loss is the absolute value of the difference of their log-likelihoods, and the threshold is set such that their probabilities must be within a factor of $1.5$, that is, $\delta=\log 1.5 $.\footnote{This task does not specify that these two pronouns should be high probability relative to other pronouns, just that they be balanced relative to each other.}
For hard negatives, we generate contexts in which a pronoun has already been used to refer to a person (presumably pronouns the person uses), and models are tested on being able to select a consistent pronoun later.

\paragraph{Temporal Entities.}
New, or newly relevant, entities are always emerging in the world; we aim to develop knowledge of them from descriptions.
We make a list of entities of new or changed relevance since 2019\footnote{The cutoff of OpenWebText \citep{Gokaslan2019OpenWeb}, which is what the Backpack of \cite{hewitt2023backpack} was trained on.} manually with the assistance of GPT-4 (prompt in Appendix~\ref{appendix_model_prompts}).
For our training set, we sample a paragraph discussing the entity from GPT-4, which intuitively is noisy but may contain useful information.
For our evaluation set, we take prefixes from the entity's Wikipedia first paragraph, and suffixes as named entities from that paragraph (Appendix~\ref{appendix_model_prompts}.)
We use negative log-likelihood loss, and set a $5\%$ probability threshold, that is, $\delta=-\log 0.05$.
Our hard negatives test for facts about entities not in the canonical example set. 

\paragraph{Hard Syntax.}
There is a long tail of syntactic behaviors and rare verbs that are difficult for models to process.
We develop a dataset based on the findings of \cite{newman2021refining}, taking rare verbs that are often misconjugated. 
For our canonical example set, we use simple agreement templates of the form \textit{$\vx$: The [singular or plural noun], $\vy^A$: [correct conjugation][suffix], $\vy^B$: [incorrect conjugation][suffix]}.
Our evaluation set uses more complex syntactic constructions with the same set of verbs, expanded from \cite{marvin-linzen-2018-targeted}.
Our loss is the difference in log-likelihoods between the correct and incorrect continuations, and our threshold requires 16x the probability on the correct conjugation suffix, that is, $\delta=\log 16$.
Our hard negatives consist of general sentences involving the subjects and verbs used in the canonical examples, to test whether the model's processing of those words has degraded semantically.

\definecolor{nicepurple}{HTML}{000000}

\section{Evaluating Finetuning Methods on Pythia LMs}

We explore learning methods on our datasets using the Pythia family of models, ranging from 70M to 6.9B parameters.
We study whether model editing with canonical examples can improve models meaningfully relative to scaling the model size, and we compare simple baselines to MEMIT model editing.

\subsection{Methods}

\paragraph{Full finetuning.}
We call finetuning all parameters of a language model \textit{full finetuning}.
Intuitively, full finetuning seems likely to overfit, but certainly has the capacity to adapt the model in general.
\begin{align}
\min_\theta \mathbb{E}_T\left[\mathcal{L}(\vx, \vy^A,\vy^B)\right]
\end{align}

Early experiments showed regularizing the learning process through KL divergence minimization with $p_{\theta_0}$ to be useful, so we use it in all finetuning-based methods (including LoRA and sense finetuning, below).
Let $R=\{\vx\}$ be a dataset of text drawn from a general corpus (and not the set $G$ used for evaluation of membership in degradation balls.).
For $\lambda \in (0,\infty)$, we approximate
\begin{align}
\min \mathbb{E}_T\left[\mathcal{L}(\vx, \vy^A,\vy^B)\right] + \lambda \mathbb{E}_{R}\left[D_\text{KL}\left(p_\theta(\cdot \mid \vx) \parallel p_{\theta_0}(\cdot \mid \vx)\right)\right].
\end{align}

\paragraph{LoRA finetuning.}
Low-Rank Adapter finetuning \citep{hu2022lora} tunes, for a set of specified matrices in $\theta$, a low-rank difference $QR$. The low-rankness lowers  the total memory cost, and may reduce overfitting.
For a set of matrices $M_1,\dots,M_k\subseteq \theta$, the updated matrices are $\{M_j+Q_jR_j\}_{j=1}^k$.
\begin{align}
\min_{\{Q_j,R_j\}_{j=1}^k} \mathbb{E}_T\left[\mathcal{L}(\vx, \vy^A,\vy^B)\right]
\end{align}
In all cases, we set the down-projection and up-projection matrices of the MLPs of the Transformer as LoRA's target matrices \citep{geva2021transformer}; we vary affected layers as a hyperparameter.

\paragraph{MEMIT.}
Mass Editing Memory in a Transformer, or MEMIT, is a state-of-the-art model editing method that targets the same MLP parameters as we've chosen for LoRA above \citep{meng2022mass}.
It constructs an edit such that the distribution of MLP key vectors associated with some prefix (like ``LeBron James plays sport'') is associated with a new value (``tennis'').
{In particular, given an association $(s_i, r_i, o_i)$, MEMIT considers the representation $h_i^L$ for the last token of $s_i$ at a target layer $L$. Via gradient descent, it computes a vector $z_i=h_i^L+d_i$ that, if used in place of $h_i^L$, would minimize the negative log-likelihood of predicting $o_i$:}
\begin{align}
z_i = h_i^L+ \argmin_{d_i}\frac1P\sum_{j=1}^P -\log p'_\theta (o_i | x_j \oplus p(s_i, r_i))\label{eqn_memit}
\end{align}
{where $p'_\theta$ indicates the distribution when substituting $h_i^L+d_i$ for $h_i^L$, and
$x_j \oplus p(s_i, r_i)$ is a prompt capturing association $i$ with random prefix $x_j$ to aid generalization.
MEMIT then spreads this update across a range of critical layers such that that $h_i^L$ approaches $z_i$. See Section 4.3 of \citet{meng2022mass} for details.}

To use MEMIT, we format our canonical examples in one of two settings.
First, we format examples so that MEMIT receives the same string-only supervision as other methods:
the subject $s_i$ is $\vx$, and the object $o_i$ is, e.g., $\vy^A$.
Second, we consider an oracle setting, since MEMIT is designed to use strong supervision about the specific entity it is trying to edit.
Here, we specify the subject of $\vx$ (underlined): ``\textit{The \underline{CEO of Renault} is Luca de Meo}''. Exact formats for each dataset are listed in Appendix \ref{appendix_memit_formats}.

By default, the negative log-likelihood in Eqn~\ref{eqn_memit} is equivalent to the the loss $\mathcal{L}$ for the country, company, and temporal datasets. For the other datasets, we modify Eqn~\ref{eqn_memit} to match the $\mathcal{L}$  in Table \ref{table_task_examples} (see Appendix \ref{appendix_memit_mods}).

\subsection{Experiments \& Results}

\paragraph{Models and Data.}
We consider Pythia models \citep{biderman2023pythia}: autoregressive Transformer language models trained on the Pile, each for 300B tokens. The model sizes we consider are 70M, 160M, 410M, 1B, 1.4B, 2.8B, and 6.9B parameters.
Apart from our canonical examples data, we use separate portions of the OpenWebText dataset \citep{Gokaslan2019OpenWeb} for our regularization set $R$ and the general corpus $G$ used to determine membership in the degradation balls.

\paragraph{Evaluation setting and hyperparameter search.}
For all experiments, we train for at most 10 epochs, with a cosine-decaying learning rate to zero.
We use a non-standard experimental setup in which hyperparameters are chosen using a \textit{validation} $(T, E)$ train and evaluation set pair, but test numbers are generated by using the best validation hyperparameters on an entirely separate (but equal-sized) \textit{test} $(T,E)$.
Recall that models must stay within a degradation ball $B_\epsilon$.
For model selection, we enforce this by training models in epochs, choosing the final epoch wherein the model is still a member of $B_\epsilon$ (or the epoch chosen by the same method at validation time, whichever is earlier.)
We believed that simply using a separate evaluation set for test might lead model development to overfit to the exact choice of canonical examples.

In early experiments, we found all methods to be highly sensitive to, e.g., the right choice of learning rate, in order to stay within the degradation balls $B_\epsilon$. %
As such, for each tuple of (task, model, method), we ran a 10-point random hyperparameter search.
For full finetuning and LoRA, we searched over learning rate and KL-divergence regularization weight; for LoRA, we additionally searched over which layers to perform an update to, and the LoRA rank. 
For MEMIT, we searched over the clamp norm factor, covariance adjustment factor $\lambda$, and KL weight described in \citet{meng2022mass}.
The details of the search are in Appendix~\ref{appendix_hyperparameter_details}.

\begin{figure}
\centering
\includegraphics[width=.30\linewidth]{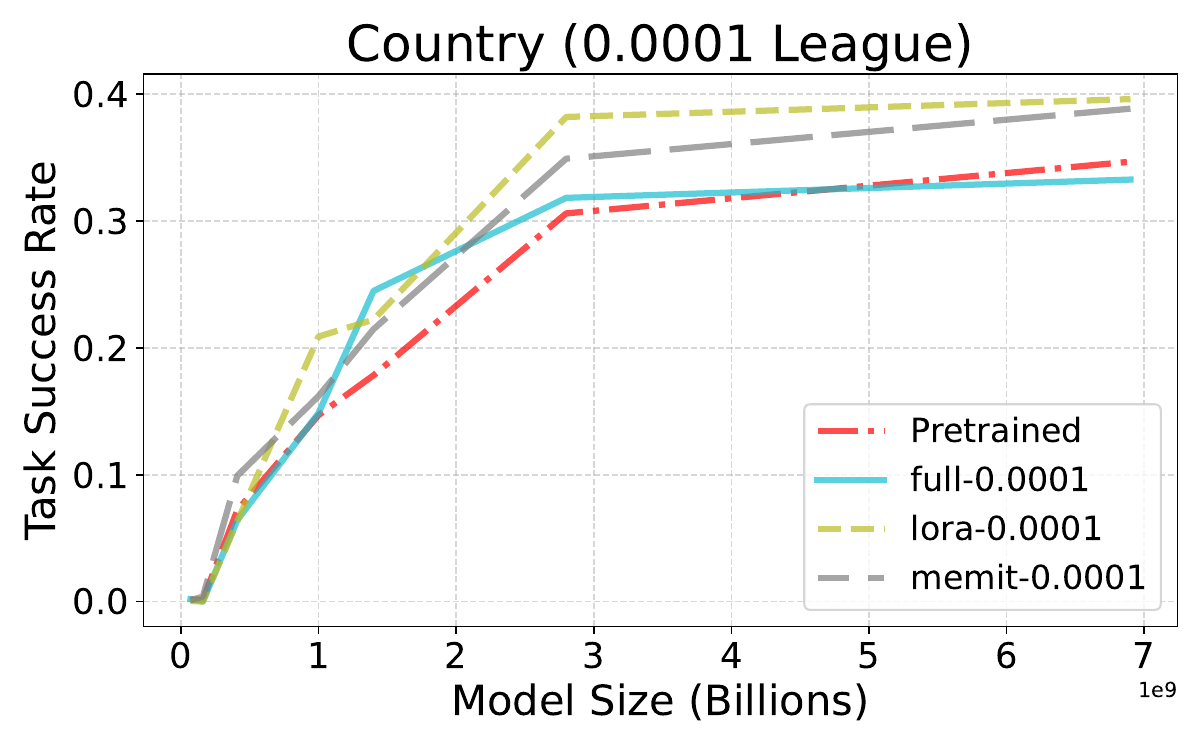}
\includegraphics[width=.30\linewidth]{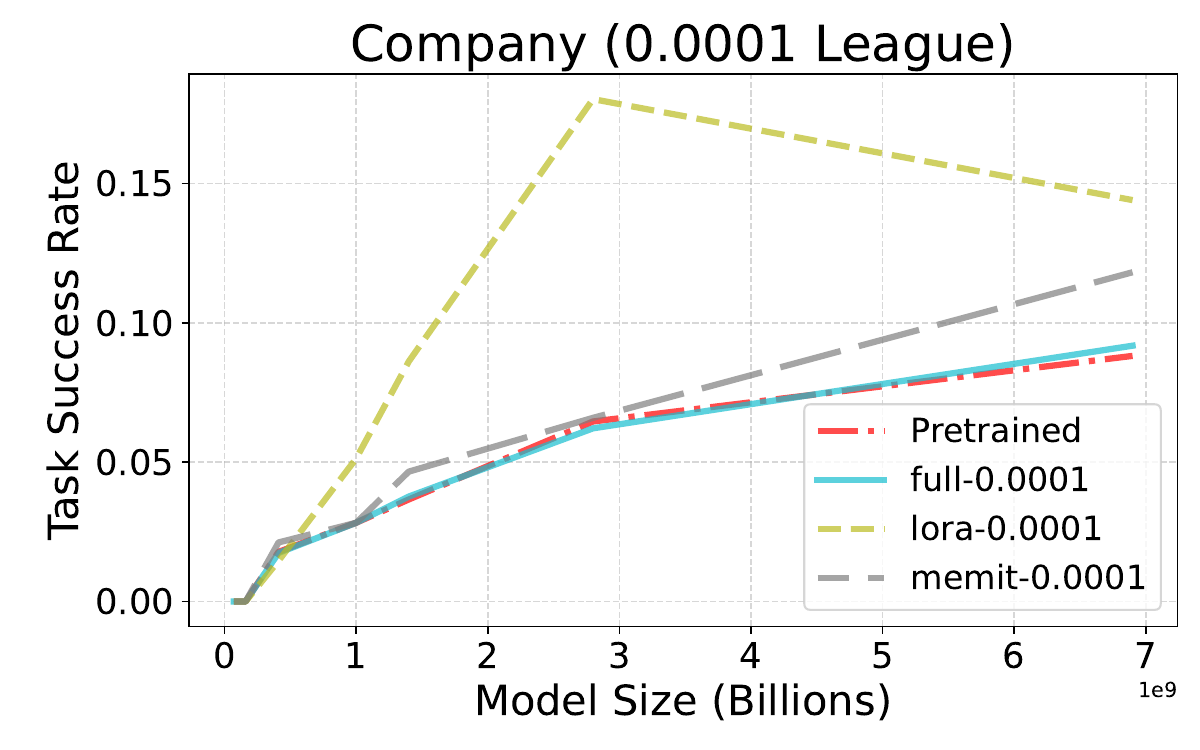}
\includegraphics[width=.30\linewidth]{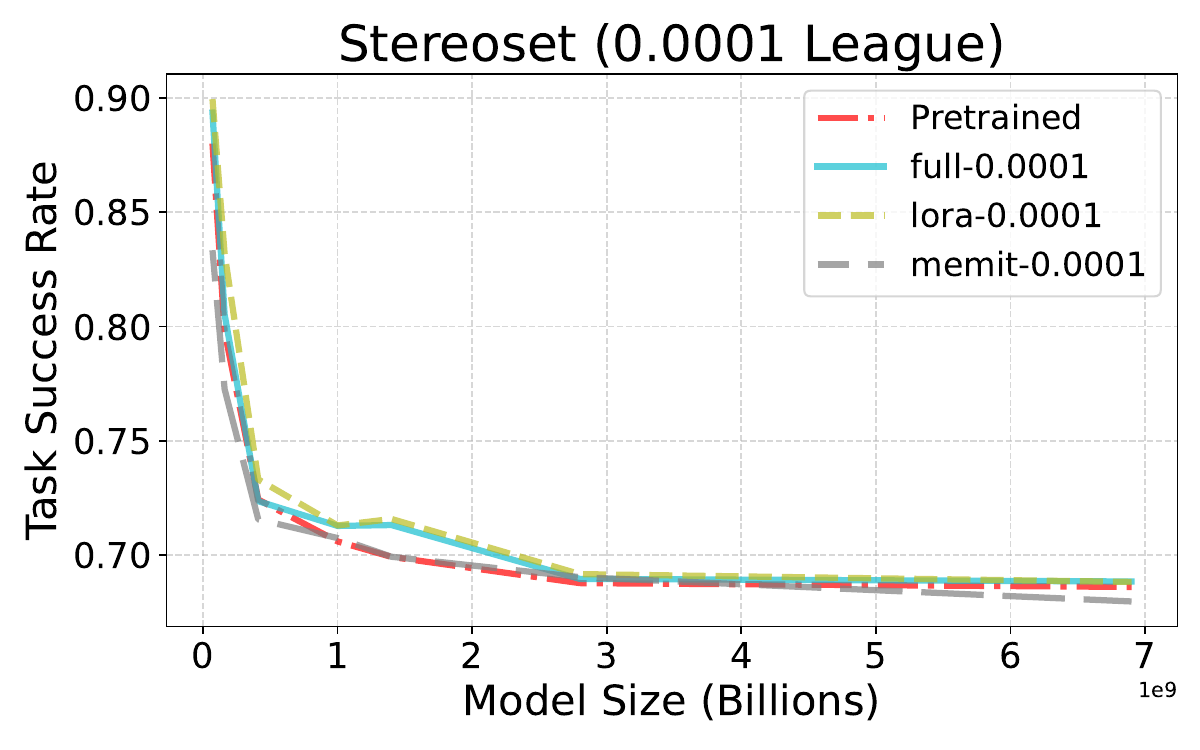}
\includegraphics[width=.30\linewidth]{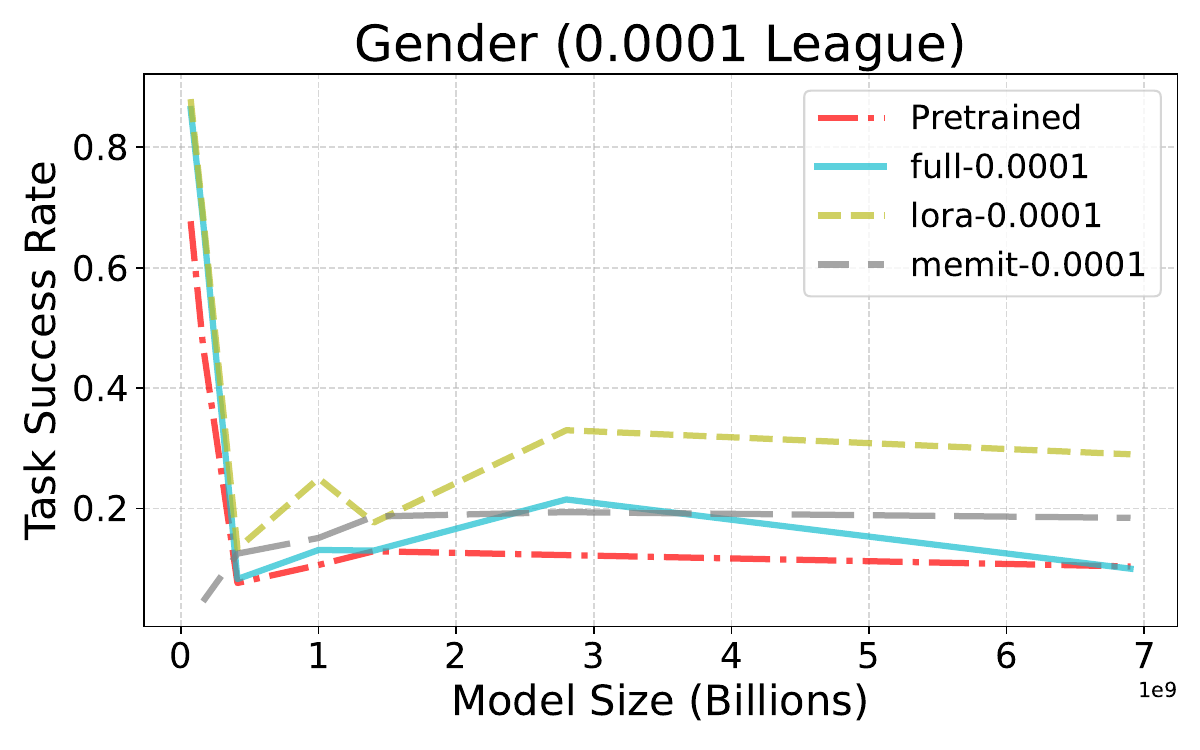}
\includegraphics[width=.30\linewidth]{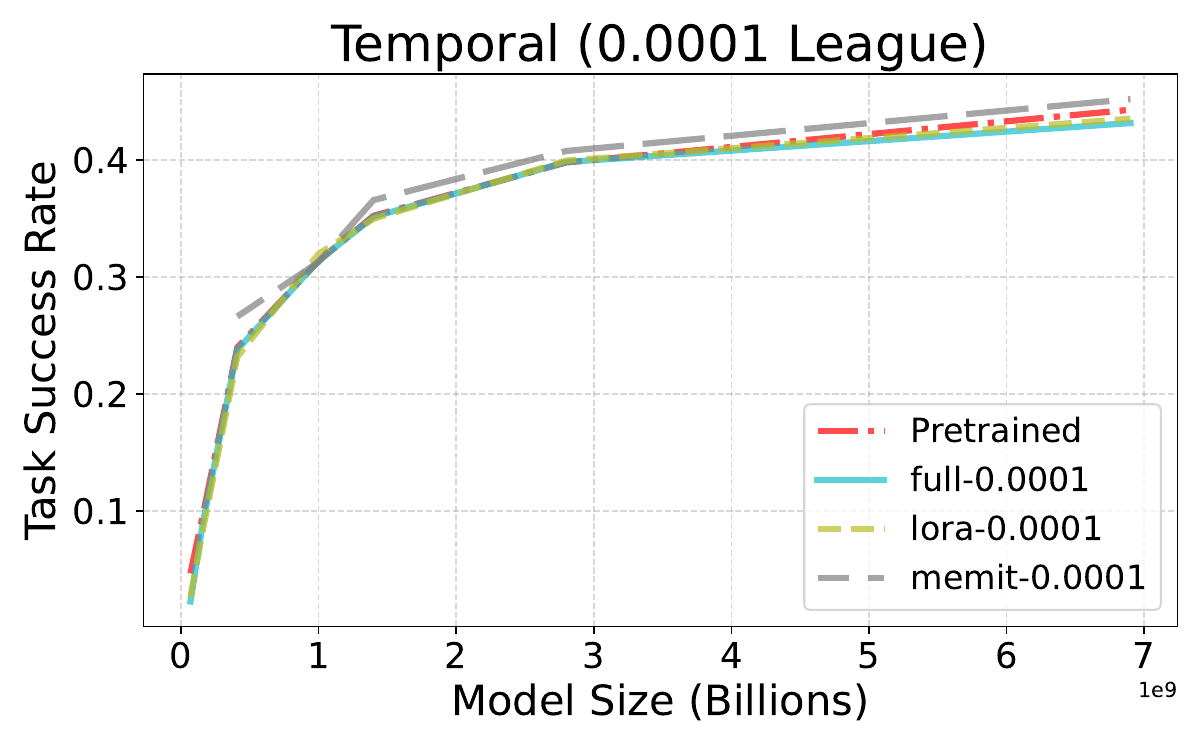}
\includegraphics[width=.30\linewidth]{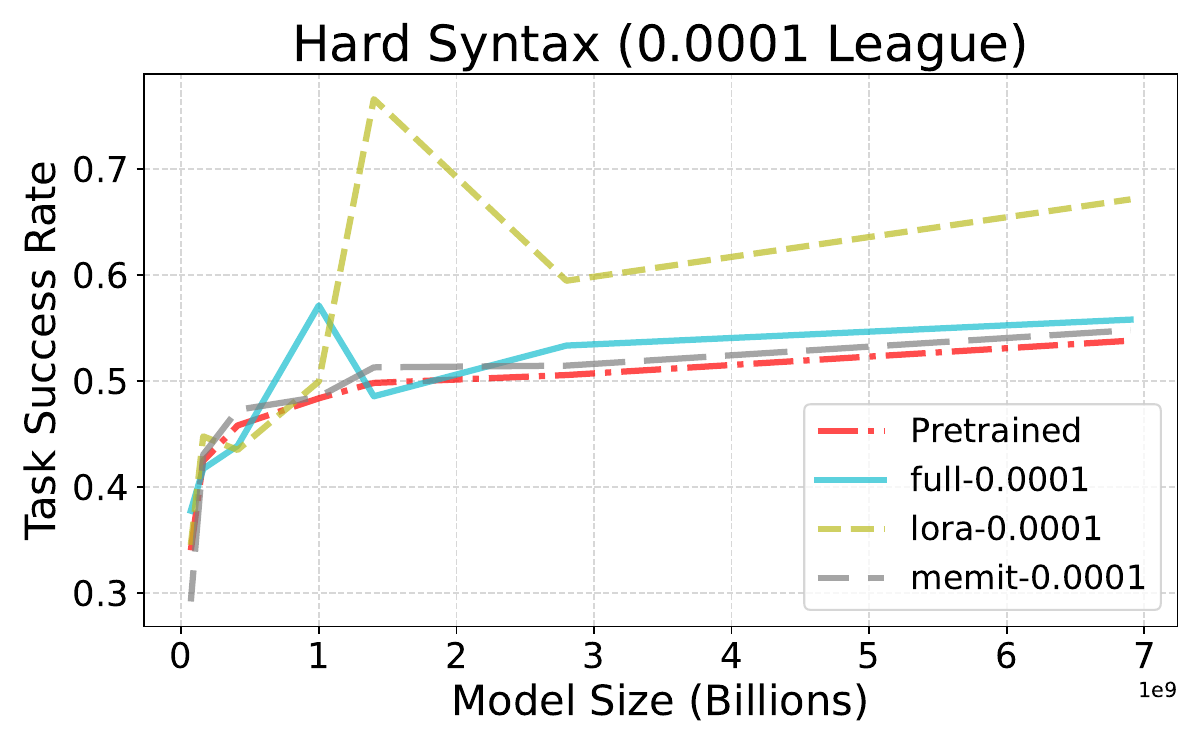}
\caption{\label{figure_pythia_results} Results for model editing with canonical examples with Pythia models for the  $B_{0.0001}$ degradation ball. Some tasks (e.g., hard syntax) show substantial improvement; others (e.g., temporal) do not.}
\end{figure}

{

\begin{wrapfigure}{R}{0.30\linewidth}
\includegraphics[width=\linewidth]{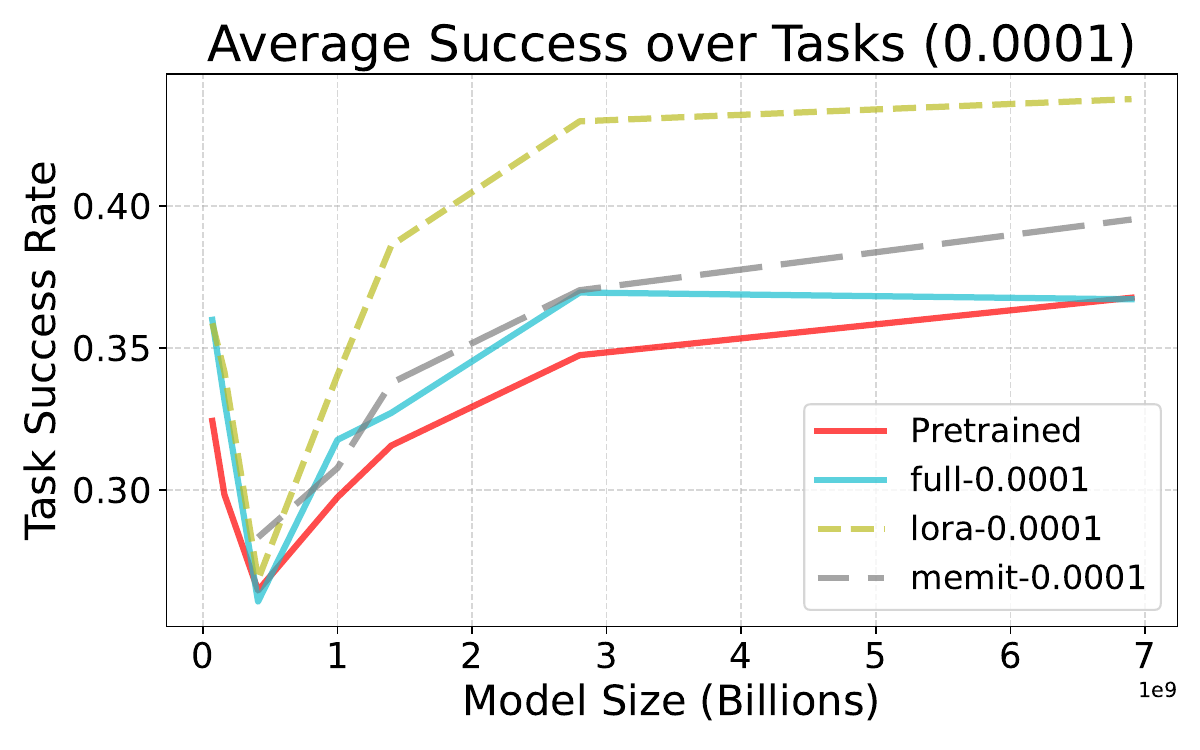}
\caption{\label{figure_pythia_avg_results}On average, LoRA outperforms other methods for model editing with canonical examples.}
\end{wrapfigure}

\paragraph{Results.}

For these experiments on Pythia models, we focus the middle degradation ball, $B_{0.0001}$, indicating that all models achieve loss on $G$ no more than a $1.0001$ factor greater than the initial model.
We find that LoRA is the strongest of the three learning methods, largely consistently across model sizes (Figure~\ref{figure_pythia_results}).
Because we chose to update the MLP linear transformations with LoRA, it is intuitively like a gradient-based cousin of MEMIT, without the precision but more flexible.
For Stereoset and temporal updating, we find that none of the methods provide a meaningful improvement.
Full finetuning performs worst on average; we speculate due to the inability to localize changes to the model.
Hard negative results are in Figure~\ref{figure_pythia_hard_neg_results}; for gender debiasing, LoRA incurs a large cost in hard negatives, and overall, MEMIT has the lowest hard negative cost.
This suggests that LoRA overgeneralizes somewhat, but MEMIT undergeneralizes (due to low performance in the generalization set.)

}

Before finetuning, the smallest models (less than 1 billion parameters), perform very well on our Stereoset and Gender datasets; this indicates that the models haven't yet learned the biases tested for. %
Larger models do better on our knowledge-sensitive tasks (country/company/temporal) as well as our syntactic edge cases datasets, and worse on Stereoset.
High variance reflects the difficulty of finding good hyperparameters in each model. Test success rates are averaged across 10 seeds.

{

\subsection{MEMIT with Oracle Supervision}

\begin{wraptable}{R}{0.3\linewidth}
\small
\centering
\raisebox{0pt}[\dimexpr\height-3\baselineskip\relax]{%
\begin{tabular}{l c c}
\toprule
& \multicolumn{2}{c}{MEMIT (0.0001)}\\
Task & Standard & Oracle\\
\midrule
Country & 2.7 & \textit{21.0} \\
Company & 1.7 & \textit{21.8} \\
Stereoset & -0.1 & \textit{0.8} \\
Hard Syntax & 1.2 & \textit{-0.2} \\
Gender & 7.3 & \textit{32.2} \\
Temporal & -0.1 & \textit{-} \\
\bottomrule
\end{tabular}
}%
\caption{\label{table_memit_oracle} Comparison of MEMIT with the standard prefix/suffix supervision compared to oracle span-level supervision.
Change in task success rate for $B_{0.0001}$ for Pythia 6.9b.
}
\end{wraptable}

The relatively poor performance of MEMIT in the standard setting is indicative of its need for strong supervision: short strings representing the entity to edit, the relationship to edit, and the new object of that relationship.
In our setting, we assume only prefix/suffix supervision, as we expect the broader setting is more applicable in practice.
However, sometimes one \textit{does} have strong supervision, and in those cases, one may want to use MEMIT.
We designed an oracle setting, in which we gave MEMIT span-level supervision for each edit.
Our results are in Table~\ref{table_memit_oracle}.
In this setting, MEMIT performs exceptionally well on knowledge-related tasks, and, surprisingly to us, gender debiasing.
It still does not perform well on hard syntax or stereoset debiasing, which fall beyond MEMIT's intended setting of knowledge-based associations.

\section{Sense Finetuning with Backpacks}
}

The Backpack was proposed as a drop-in replacement for the Transformer that provides a reliable interface for intervention in the network, to allow for interpretability and control \citep{hewitt2023backpack}.
In this section, we briefly review the Backpack, and present \textit{sense finetuning}, a new finetuning method for the Backpack that automates interpretability work and performs well for model editing with canonical examples.

\subsection{The Backpack Language Model}
The Backpack language model learns a set of $k$ word2vec-like sense vectors $c(x)_\ell\in\mathbb{R}^{d}$ for each element of the vocabulary $x\in\mathcal{V}$, where $d$ is the model's common vector dimensionality.
To construct a distribution, the Backpack weights and sums the sense vectors of the words in the prefix:
\begin{align}
 &p_\theta(\cdot \mid \vx_{1:t}) = \text{softmax}(Eh_{t})\\[13pt]
 &h_t = \sum_{j=1}^t \sum_{\ell=1}^k \eqnmarkbox[blue]{a1}{\vc(x_j)_\ell} \eqnmarkbox[red]{a2}{\alpha_{tj\ell}(\vx_{1:t})}
\end{align}
\annotate[yshift=-.75em]{below, label below}{a2}{Weighting of sense in prediction}%
\annotate[yshift=0.3em]{above, label below}{a1}{Sense vector $\ell$ of word $j$, an $\mathbb{R}^d$ word2vec-like word vector}%
\vspace{5pt}

\noindent
where $E\in\mathbb{R}^{|\mathcal{V}|\times d}$ is the softmax matrix, and $\alpha\in\mathbb{R}^{n\times n\times \ell}$ is a matrix of non-negative, autoregressively masked weights. %
The expressivity of the Backpack comes from its construction of the $\alpha$ function, which for the model of \cite{hewitt2023backpack}, is a Transformer.
Despite this expressivity, the final prediction is still a weighted sum over the 
sense vectors $c(x_j)_\ell$.
\cite{hewitt2023backpack} found that the sense vectors of words specialize unsupervisedly during the language model training process to encode rich aspects of language use.%

\begin{figure}
\includegraphics[width=\linewidth]{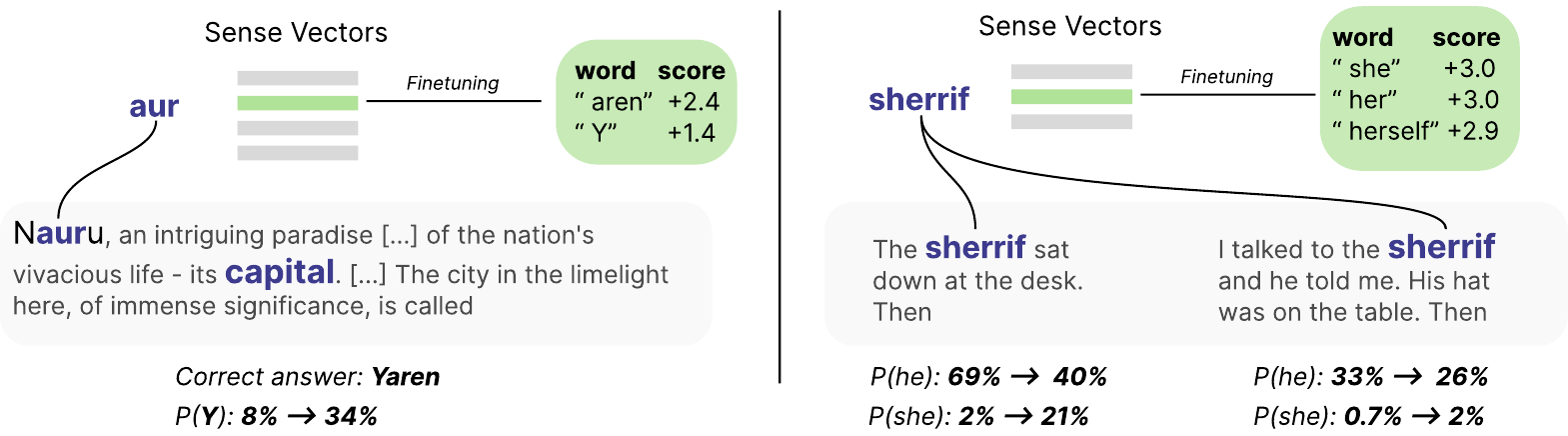}
\caption{\label{figure_sense_ft_viz}
In sense finetuning, a handful of sense vectors are selected based on an estimate of their importance to the canonical example relative to general text. In one example, a subword \texttt{aur} of the name of the country Nauru has some of its sense vectors finetuned. Finetuning updates the sense vector to, in this case, store knowledge about the capital of the country.}
\end{figure}

\subsection{Sense Finetuning}
In \cite{hewitt2023backpack}, the authors hand-pick a few sense vectors that seem to represent a concept, and manually specify transformations to edit them to make changes to the language model.
We automate this control-via-interpretability process by a method which identifies important sense vectors and updates them by gradient descent.\footnote{The specific parameterization of the Backpack shares weights in the sense vectors by generating them by a common feed-forward network that takes word embeddings as input. This was done to reduce the total parameter count, since independently parameterizing all $k|\mathcal{V}|=804112$ vectors (at $768$ parameters per vector) would require 620M parameters, significantly more than the 124M used to define the Transformer-based weight network. The shared parameterization takes 46M. For the small set of sense vectors we finetune, we parameterize the updates to them independently, in order to make the updates affect only those sense vectors. This adds a small number of extra learnable parameters to the network.}

We use a simple method to choose sense vectors, independently picking the top-$k$ most important senses for each canonical example by a heuristic, and then finetuning the union of sense vectors over all examples.
Most parameters of the network (including all that participate in the contextualization $\alpha$) are frozen.
For a target token $\vy_t^A$, let $\alpha_{tc}$ be the weight assigned to sense vector $c\in C$ in predicting $\vy_t^A$.
We score each sense vector $c$ for a single example as:
\begin{align}
\text{importance}(c; \vx, \vy^A,\vy^B) =\sum_{t=1}^{|\vy^A|} \alpha_{tc}+ \sum_{t=1}^{|\vy^B|} \alpha_{tc} - \lambda \mathbb{E}_R[\sum_{t=1}^{|\vx|}\alpha_{tc}]. \label{eqn_sense_selection}
\end{align}
That is, we take senses that are weighted more under the canonical example than under the regularization distribution.
Figure~\ref{figure_sense_ft_viz} visualizes senses chosen and finetuned for our tasks.

\subsection{What sense finetuning teaches: a look at the gradient}

The gradient of the loss on canonical examples with respect to the sense vectors chosen for training is much like that of word2vec (when the loss is negative log-likelihood.)
In particular, due to linearity, the senses are simply updated to point more in the directions of the word embeddings of target words; the strength of their update depends on $\alpha$, the weight they are assigned in the Backpack sum:
\vspace{10pt}
\begin{align}
\nabla_{c} \mathbb{E}_T\left[\mathcal{L}(\vx, \vy^A, \vy^B)\right] &= -\mathbb{E}_T\left[ \sum_{t=1}^{|\vy^A|} \eqnmarkbox[blue]{a1}{\alpha_{tc}}( \eqnmarkbox[red]{a2}{E_{\vy^A_t}} - \eqnmarkbox[black]{a3}{\sum_{w\in\mathcal{V}} p_\theta(w\mid \vx, \vy_{1:t-1}) E_w})\right].
\end{align}
\annotate[yshift=1em]{above, label below}{a1}{Weight to which the sense is incorporated into prediction}
\annotate[yshift=-3em]{below, label below}{a2}{Embedding of true next word}
\annotate[yshift=-0.5em]{below, label below}{a3}{Average predicted embedding}\\

\vspace{10pt}

\noindent
Hence, due to sense vectors combining log-linearly for prediction, \textit{whenever} these updated senses are assigned high $\alpha$ by the Backpack at inference time, the effect of finetuning is the same: to increase the score of the words in the canonical example.%

\begin{table}
\centering
\definecolor{nicepurple}{HTML}{593C8F}
\small
\begin{tabular}{lrrrrrrrrrrrrr}
  \toprule
  Task& Initial & \multicolumn{3}{c}{$\Delta$, $B_{0.001}$ $\uparrow$} & & \multicolumn{3}{c}{ $\Delta$,  $B_{0.0001}$ $\uparrow$ } & & \multicolumn{3}{c}{$\Delta$, $B_{10^{-5}}$ $\uparrow$} \\
  \cmidrule(r){3-5} \cmidrule(r){7-9} \cmidrule{11-13}
   & & \text{Full} & \text{LoRA} & \text{Senses} & & \text{Full} & \text{LoRA} & \text{Senses} & & \text{Full} & \text{LoRA} & \text{Senses} \\
  \cmidrule(r){3-5} \cmidrule(r){7-9} \cmidrule{11-13}
  \text{Stereoset}& 76.3&1.1&0.9&\color{nicepurple}\bf7.8& &0.3&0.1&\color{nicepurple}\bf3.8& &0.0&0.0&\color{nicepurple}\bf1.9&  \\
\text{Country}& 9.9&4.9&3.4&\color{nicepurple}\bf8.2& &2.3&1.5&\color{nicepurple}\bf4.3& &2.0&1.7&\color{nicepurple}\bf2.6&  \\
\text{Company}& 3.1&\bf5.3&0.4&\color{nicepurple}4.9& &0.4&0.3&\color{nicepurple}\bf0.6& &0.2&-0.2&\color{nicepurple}\bf1.6&  \\
\text{Gender}& 9.2&5.2&-0.9&\color{nicepurple}\bf13.9& &-0.6&-0.1&\color{nicepurple}\bf11.7& &-0.5&-0.8&\color{nicepurple}\bf12.0&  \\
\text{Hard Syntax}& 56.4&\bf16.7&15.7&\color{nicepurple}16.4& &2.4&1.1&\color{nicepurple}\bf15.1& &0.0&0.0&\color{nicepurple}\bf10.6&  \\
\text{Temporal}& 23.0&\bf1.1&0.7&\color{nicepurple}0.5& &0.3&\bf0.8&\color{nicepurple}0.6& &0.2&0.1&\color{nicepurple}\bf0.2&  \\
\midrule
\text{Average}& 29.6& 5.7& 3.4& \color{nicepurple} \bf 8.6& & 0.8& 0.6& \color{nicepurple} \bf 6.0& & 0.3& 0.1& \color{nicepurple} \bf 4.8&  \\
  \bottomrule
  \end{tabular}
\caption{\label{table_backpack_results} 
Comparison of success rate improvements on model editing with canonical examples at three degradation balls for full finetuning, LoRA, and sense finetuning on the Backpack. Sense finetuning substantially outperforms other methods.}
\end{table}

\subsection{Experiments \& Results}

We now evaluate whether our sense finetuning improves over full finetuning, LoRA, and MEMIT for the 170M parameter Backpack language model trained by \cite{hewitt2023backpack}.

\paragraph{Hyperparameter search.}

In addition to learning rate and KL-divergence regularization, we have new hyperparameters $k$ (number of senses to finetune) and regularization weight in sense selection.
For all methods, for all tasks, we sample 25 configurations in our hyperparameter search, picking the best method to train and evaluate on our test settings.
All other experimental choices are the same as for the Pythia experiments.

\paragraph{Results.}

We find that across degradation balls, sense finetuning performs best in generalization out of all methods.
It is especially strong, however, in the more stringent $B_{0.0001}$ and $B_{10^{-5}}$ degradation balls, which allow little deviation from the original language model.
On hard negatives, we find that LoRA and full finetuning incur almost no degradation. Sense finetuning incurs more degradation, indicating some overgeneralization, except in $B_{10^{-5}}$, where it too achieves close to zero degradation.
We find that sense finetuning is particularly strong for de-stereotyping (both for Stereoset and gender bias).
Our results for generalization are in Table~\ref{table_backpack_results}, and results for hard negatives in Table~\ref{table_backpack_hard}.

\section{Improving LLMs with Sense Finetuned Backpacks}
Given a large pretrained model (not a Backpack), we now show how we can improve it using sense finetuning.
We sense finetune a small Backpack and then ensemble the capabilities of the large model with the improvements of the sense finetuning using an inference-time ensemble \cite{liu2021dexperts,mitchel2023emulator}.

\paragraph{Method.}
Let $p_\text{large}$ be a large language model that we would like to improve with canonical examples.
We cannot improve it via sense finetuning because it does not in general have sense vectors.
Let $p_\text{bp}^{\text{pre}}$ be a pretrained language model (ours will be a Backpack), and $p_\text{bp}^{\text{ft}}$ be a language model finetuned on canonical examples.
Intuitively, we want to impart the adaptations of the canonical example finetuning to a larger language model $p_\text{large}$.
We do so by the following:
\begin{align}
\log p_\text{large}^{\text{ft}} \propto \beta (\log p_\text{bp}^{\text{ft}}  - \log p_\text{bp}^{\text{pre}}) + \log p_\text{large}^{\text{pre}}.
\end{align}
Intuitively, since the pretrained and finetuned Backpacks are within $\epsilon$ loss of each other, adding their difference of logits should only rarely make large changes to $p_\text{large}$.\footnote{We run a coarse search (in increments of $0.1$) for a value of $\beta$ as close to $1$ as possible while ensuring the resulting model is in the correct degradation ball.} 
This simple heuristic recently used in the setting of approximating finetuning large models by finetuning small models, by \citet{mitchel2023emulator}.

\paragraph{Experiments \& Results}
We use the GPT-J-6B model \citep{gpt-j}, comparing full finetuning and LoRA finetuning to our proposed ensemble.
We choose GPT-J since it uses the same tokenization as our Backpack.
We do no further finetuning of the GPT-J model in the ensemble.\footnote{Running both Backpacks takes only marginally more compute than running one (see
Appendix~\ref{appendix_backpack_efficiency}).}
We run a 10-point random hyperparameter sweep on the validation set for the GPT-J finetuning methods.

Generalization results are in Table~\ref{table_gptj_test}, and hard negatives results in Table~\ref{table_gptj_hard}.
We find that for the most strict degradation ball $B_{10^{-5}}$, our Backpack ensemble even substantially outperforms both finetuning methods for GPT-J in generalization, at no cost in hard negative performance.
For the less strict degradation balls, our ensemble performs slightly worse than the other methods.
This result is evidence that the Backpack with sense tuning is \textit{more adaptable} than the 35x-larger GPT-J, and with our ensemble, we can impart the benefits of these adaptations to the larger model.

\begin{table}
\definecolor{nicepurple}{HTML}{593C8F}
\centering
\small
  \begin{tabular}{lrrrrrrrrrrrrr}
  \toprule
  Task & \text{Initial} & \multicolumn{3}{c}{$\Delta$, $B_{0.001}$ $\uparrow$} & & \multicolumn{3}{c}{ $\Delta$,  $B_{0.0001}$ $\uparrow$ } & & \multicolumn{3}{c}{$\Delta$, $B_{10^{-5}}$ $\uparrow$} \\
  \cmidrule(r){3-5} \cmidrule(r){7-9} \cmidrule{11-13}
  & & \text{Full} & \text{LoRA} & \text{Senses} & & \text{Full} & \text{LoRA} & \text{Senses} & & \text{Full} & \text{LoRA} & \text{Senses} \\
  \cmidrule(r){3-5} \cmidrule(r){7-9} \cmidrule{11-13}
  \text{Country}& 42.8&9.2&10.9&\color{nicepurple}\bf11.2& &3.2&\bf11.1&\color{nicepurple}6.4& &-0.1&3.5&\color{nicepurple}\bf4.2&  \\
\text{Company}& 13.6&11.6&\bf16.0&\color{nicepurple}5.1& &1.9&\bf16.6&\color{nicepurple}1.0& &0.1&0.0&\color{nicepurple}\bf2.0&  \\
\text{Stereoset}& 68.9&2.2&0.5&\color{nicepurple}\bf9.1& &0.4&0.5&\color{nicepurple}\bf4.0& &0.1&0.0&\color{nicepurple}\bf1.9&  \\
\text{Hard Syntax}& 54.5&24.2&\bf31.7&\color{nicepurple}18.7& &6.1&6.2&\color{nicepurple}\bf18.1& &-0.1&2.0&\color{nicepurple}\bf11.9&  \\
\text{Gender}& 13.6&\bf22.1&5.6&\color{nicepurple}6.1& &2.4&2.3&\color{nicepurple}\bf5.0& &0.2&0.3&\color{nicepurple}\bf4.7&  \\
\text{Temporal}& 47.8&-0.3&\bf-0.0&\color{nicepurple}-0.7& &-0.4&\bf-0.3&\color{nicepurple}-0.6& &-0.4&\bf0.4&\color{nicepurple}0.0&  \\
\midrule
\text{Average}& 40.2&\bf 11.5& 10.8& 8.3& & 2.3& \bf 6.1& 5.6& & -0.0& 1.0& \bf \color{nicepurple} 4.1&  \\

  \bottomrule
  \end{tabular}
 \vspace{4pt} 
  \caption{\label{table_gptj_test}
Comparison of success rate improvements on model editing with canonical examples at three degradation balls for full finetuning, LoRA, and the sense finetuned Backpack ensemble for GPT-J. For the most conservative degradation ball, our Backpack methods outperforms the other methods.}
\end{table}

\subsection{Visualizing Backpack improvements}
To provide intuition for how sense finetuning updates a model, we provide two examples in Figure~\ref{figure_sense_ft_viz}.
The first canonical example is \textit{The capital of Nauru is Yaren}.
Because of their greater importance to the canonical example than to general text (Eqn~\ref{eqn_sense_selection}), sense vectors of the subword \texttt{aur} in \textit{Nauru} are chosen for finetuning.
The result of finetuning is to increase the score of the subwords of Yaren, \texttt{Y} and \texttt{aren}, under the sense vector---this score is not dependent on context, and contributes additively to the model predictions with weight $\alpha$.
Thus, when the network chooses to look at the finetuned senses, it will \textbf{always} score the corresponding words more highly relative to the pretrained model.
Thus, changing lexical associations are the most obvious uses for sense finetuning.
In the canonical example \textit{The sheriff said \{he, she\}}, sense vectors of \textit{sheriff} are finetuned to score words like \textit{her} more highly---but note that when an explicit pronoun is used in context, the model can still copy from the prior pronoun.

\section{Discussion \& Conclusion}
In this work, we presented \textit{model editing with canonical examples}, a problem setting that centers learning from a single example, evaluating out-of-distribution, and strictly limiting deviation from the original model.
We've found that simple finetuning methods like LoRA can improve models somewhat with canonical examples while keeping the model's loss within a factor of $1+10^{-4}$.
However, it is difficult to precisely edit models, especially since only string supervision is provided, as shown by the decrease in performance of MEMIT compared to its performance when it receives stronger supervision.
We've shown that the Backpack's sense vectors provide a useful method for model editing with canonical examples, even for improving the 35x larger GPT-J model more than finetuning GPT-J itself in one setting.
We hope that the setting of model editing with canonical examples will help spur research in understanding and robust improvement of LLMs.

The architecture of a neural model has implications not just for its computational efficiency and inductive bias, but also for the kinds of fixes we can make to it after it's trained.
The Backpack and its lexically-defined sense vectors allow for precise edits of lexical selections.
In exploring new model architectures, we suggest directly designing in components corresponding to the kinds of fixes we want to be able to make.
While it's costly to train new models with new architectures, we can leverage small, adaptable models to fix monolithic large models, like we've shown here with GPT-J.

\bibliography{main}
\bibliographystyle{tmlr}

\appendix

\section{Efficiency of running a Backpack `twice'}\label{appendix_backpack_efficiency}
In our ensemble, 
\begin{align}
\log p_\text{large} \propto \beta (\log p_\text{bp}^{\text{ft}}  - \log p_\text{bp}^{\text{pre}}) + \log p_\text{large},
\end{align}
it looks like we have to run two Backpacks: the finetuned and the pretrained models.

However, we've only finetuned the senses of the Backpack.
Referencing the Backpack contextualization function:
\begin{align}
 &p_\theta(\cdot \mid x_1,\dots,x_t) = \text{softmax}(Eh_{t})\\
 &h_t = \sum_{j=1}^t \sum_{\ell=1}^k \vc(x_j)_\ell \alpha_{tj\ell},
\end{align}
we see that the the weights of the Backpack sum $\alpha= f(x_1,\dots,x_t)$ do not change as a function of the sense vectors $\vc(x)$.
Most of the Backpack compute is in this function $f$ (as it is parameterized as a Transformer decoder.)
Hence, when computing the forward pass of a Backpack twice for our ensemble, we can cache $\alpha$, and only recompute the final sum.

\section{Hard Negatives Results} \label{appendix_hard_negatives}

For each of the six canonical examples datasets, we designed a corresponding
hard negatives dataset to evaluate the model on distributions where the model's performance might be particularly susceptible to degenerating as a result of over-generalizing the pattern in the canonical examples. Descriptions and examples for each hard negatives task are in Table~\ref{table_hard_neg_datasets}. The design of hard negatives tasks can be categorized into two types: 

\begin{enumerate}
    \item Tests whether model performance drops with respect to similar entities that did not appear in the canonical examples. (Here, for company-CEO and temporal update.)
    \item For entities that did appear in the canonical examples, tests whether the model becomes less capable of modeling other orthogonal properties of theirs. (Here, for country-capital, Stereoset, gender bias, and hard syntax.)
\end{enumerate}

To measure the degradation, we compute the negative log-likelihood assigned to the true completion $\vy$ before and after finetuning, and take the difference. Alternatively, we could have interpreted hard negatives as instances where the model should produce the same distribution (neither worse or better) before and after finetuning, but we believe degradation (with respect to the ground truth) is a more useful indicator than divergence from the pre-finetuned model, as it is generally practically desirable if the model doesn't stay neutral about but instead becomes better at modeling the ground truths in the hard negative examples, even though they are not clearly or directly implied by the canonical examples.

\begin{table}
\centering
\small
\begin{tabular}{l p{65mm} p{70mm} c}
\toprule
\bf Task & \bf Hard Negative Task & \bf Example \\
\midrule
Country & For countries in the canonical examples, predict cities other than the capital city when appropriate. The input $\vx$ mentions the country and then elicits a non-capity city by providing a factual description about this other city which is not true, or much less true, of the capital. & \textit{Japan is renowned for its preserved and maintained traditional temples, which can be seen throughout the city of} \textbf{Kyoto} \\
\midrule
Company & Predict CEOs of companies that were not in the canonical examples. & \textit{WeWork, a renowned company revolutionizing the concept of shared workspaces, has been making waves in the business world. Led by} \textbf{Sandeep Mathrani} \\
\midrule
Stereoset & For entities in the canonical examples, predict their definitions in PyDictionary. & \textit{The definition of Iraq is} \textbf{a republic in the Middle East in western Asia; the ancient civilization of Mesopotamia was in the area now known as Iraq} \\
\midrule
Gender Bias & For careers in the canonical examples, when the worker's pronoun has been explicitly indicated in the context $\vx$ and another pronoun is now elicited, predict the consistent pronoun. & \textit{With her steady hands and compassionate heart, this nurse has transformed countless lives in her career of service. Every weekday,} \textbf{she} \\
\midrule
Temporal & Predict related named entities for subjects for which facts have stopped changing five years ago (before 2019). & \textit{Galileo was an American robotic space probe that studied the planet Jupiter and its moons, as well as the asteroids} \textbf{Gaspra} \\
\midrule
Hard Syntax & Generate semantically coherent sentences about the subjects and verbs that showed up in the canonical examples. & 1. Subject: \textit{\underline{Bankers}} \textbf{work diligently to manage and invest funds for their clients while navigating the ever-changing financial landscape.} 2. Verb: \textit{Many} \textbf{individuals signed \underline{petitions} to advocate for change in their communities.} \\
\bottomrule
\end{tabular}
\vspace{4pt}
\caption{\label{table_hard_neg_datasets} Hard negative task description and example for each of our six canonical example datasets. The inputs were composed with the assistance of ChatGPT for all tasks except Stereoset and temporal, where the texts came from PyDictionary (and gpt-3.5-turbo if no dictionary entry existed) and Wikipedia respectively.}
\end{table}

\begin{figure}
\centering
\includegraphics[width=.32\linewidth]{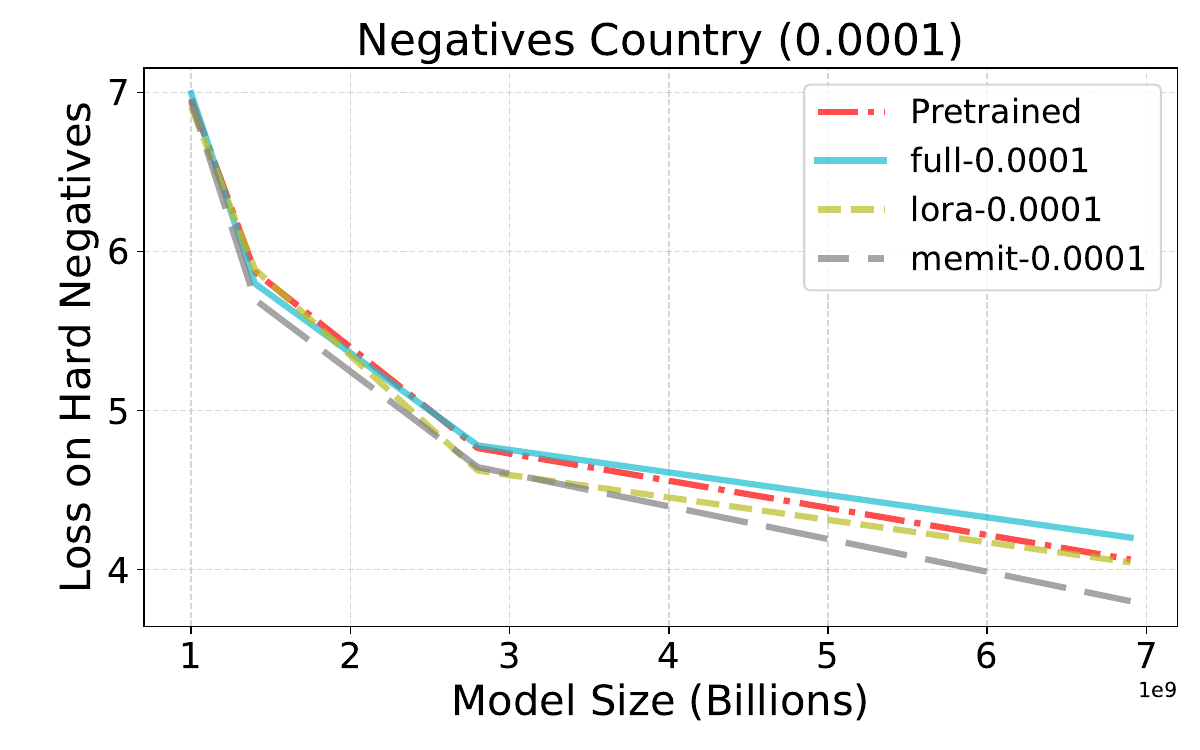}
\includegraphics[width=.32\linewidth]{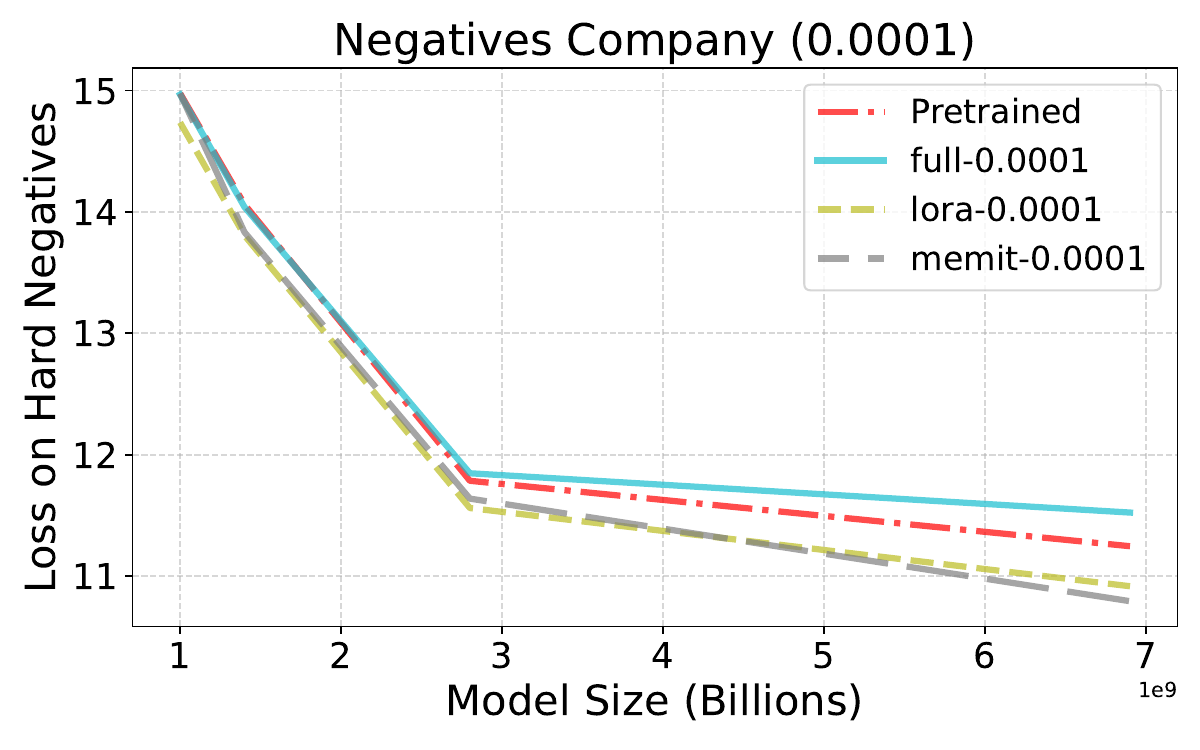}
\includegraphics[width=.32\linewidth]{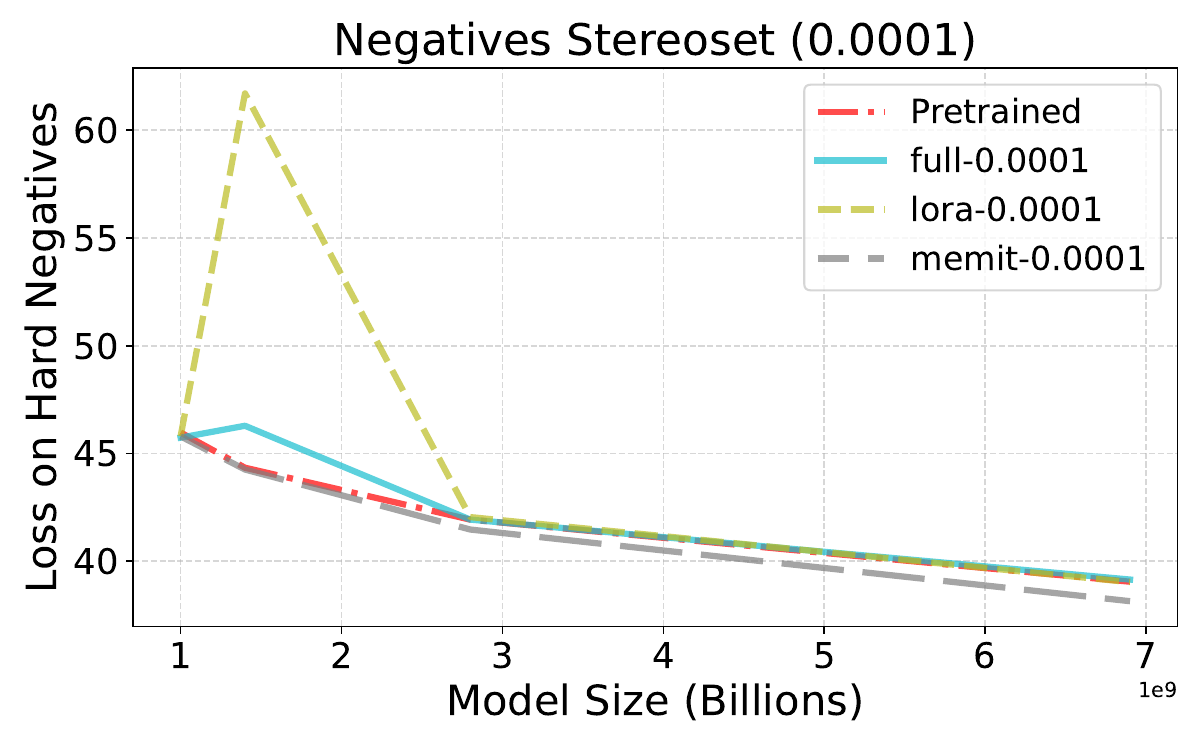}
\includegraphics[width=.32\linewidth]{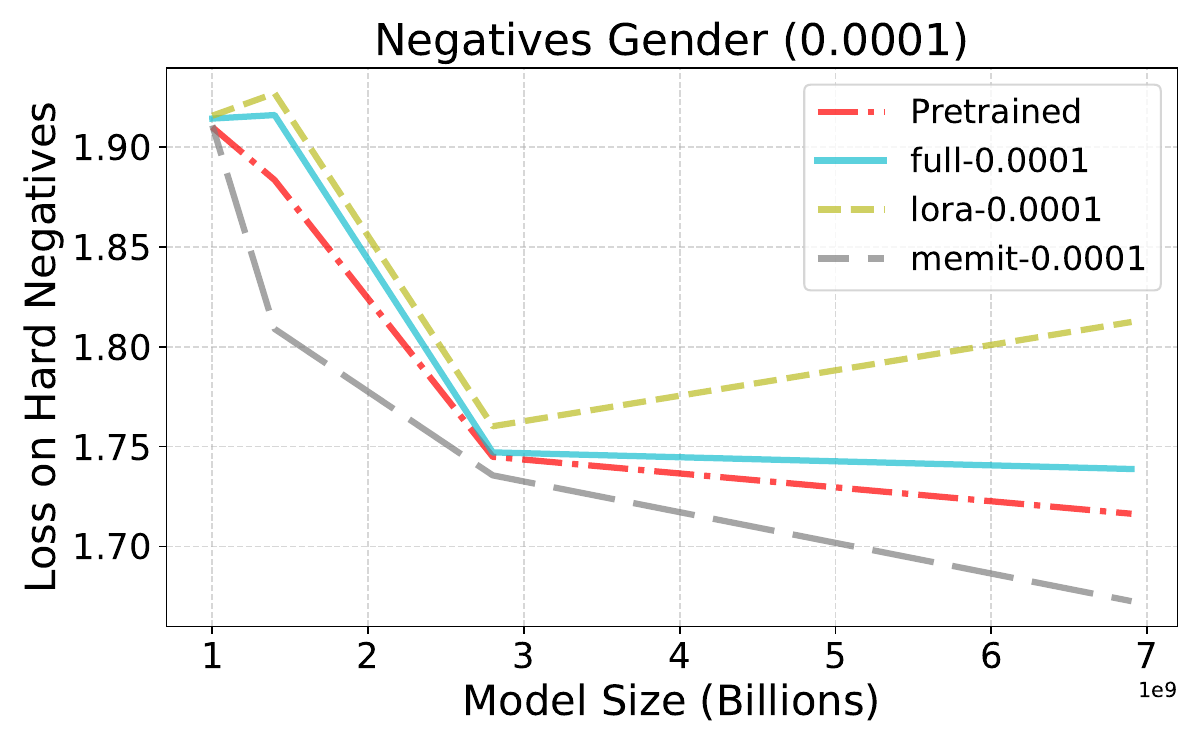}
\includegraphics[width=.32\linewidth]{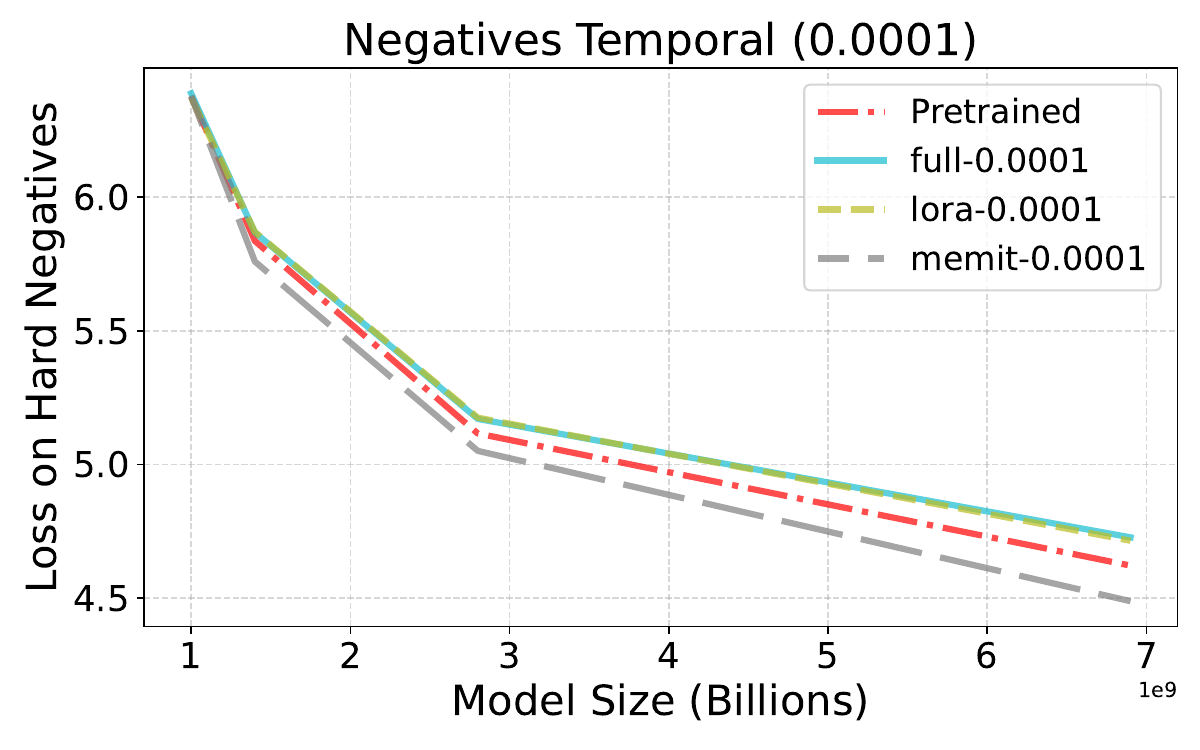}
\includegraphics[width=.32\linewidth]{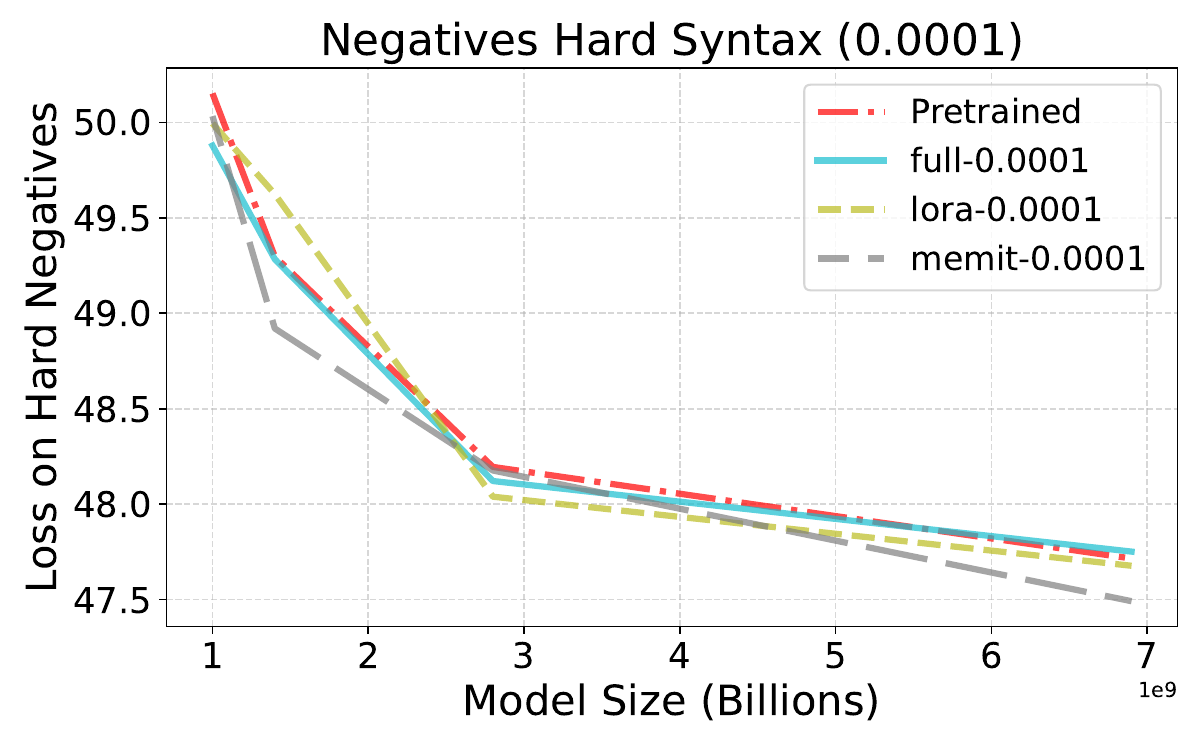}
\caption{\label{figure_pythia_hard_neg_results}Hard Negatives Results for Pythia in ball 0.001. Lower is better. Note that MEMIT improves performance slightly on hard negatives (but, as shown in Figure~\ref{figure_pythia_results}, was less effective at generalization.)}
\end{figure}

The hard negatives results are in Tables~\ref{table_backpack_hard}~and~\ref{table_gptj_hard}.
We find that sense finetuning tends to perform worse on hard negatives except in the most stringent ball $B_{10^{-5}}$ and in fact, other methods often \textit{improve} performance on hard negatives.

For Pythia models, hard negatives results are in Figure~\ref{figure_pythia_hard_neg_results}.
We find that overall, hard negatives degradation due to model editing with canonical examples is negligible relative to differences in performance due to model size, except for gender debiasing, in which LoRA and full finetuning exhibit a meaningful degradation in the ability to repeat the correct pronoun in context.
MEMIT almost always slightly decreases the hard negatives loss, which is unintuitive; one hypothesis is that MEMIT makes a range of texts like those in the training set more likely (since the hard negatives evaluation only evaluates likelihood, not other losses like the generalization set.)

\begin{table}
\definecolor{nicepurple}{HTML}{593C8F}
\small
\centering
\begin{tabular}{lrrrrrrrrrrrrr}
  \toprule
  Task & \text{Initial} & \multicolumn{3}{c}{$\Delta$, $B_{0.001}$ $\downarrow$} & & \multicolumn{3}{c}{ $\Delta$,  $B_{0.0001}$ $\downarrow$ } & & \multicolumn{3}{c}{$\Delta$, $B_{10^{-5}}$ $\downarrow$} \\
  \cmidrule(r){3-5} \cmidrule(r){7-9} \cmidrule{11-13}
  & & \text{Full} & \text{LoRA} & \text{Senses} & & \text{Full} & \text{LoRA} & \text{Senses} & & \text{Full} & \text{LoRA} & \text{Senses} \\
  \cmidrule(r){3-5} \cmidrule(r){7-9} \cmidrule{11-13}
  \text{Country}& 10.8&\bf-0.1&-0.0&\color{nicepurple}0.2& &-0.1&\bf-0.1&\color{nicepurple}-0.0& &\bf-0.2&-0.1&\color{nicepurple}-0.0&  \\
\text{Company}& 18.2&\bf-0.3&-0.2&\color{nicepurple}0.3& &\bf-0.4&-0.4&\color{nicepurple}0.0& &-0.1&\bf-0.2&\color{nicepurple}0.0&  \\
\text{Stereoset}& 51.9&\bf0.1&2.1&\color{nicepurple}7.2& &\bf0.1&0.3&\color{nicepurple}0.5& &\bf0.0&0.0&\color{nicepurple}0.0&  \\
\text{Hard Syntax}& 58.1&\bf-0.1&0.1&\color{nicepurple}5.4& &\bf-0.0&-0.0&\color{nicepurple}1.9& &\bf-0.0&-0.0&\color{nicepurple}0.1&  \\
\text{Gender}& 1.7&0.0&\bf-0.0&\color{nicepurple}0.0& &0.0&\bf0.0&\color{nicepurple}0.0& &0.0&0.0&\color{nicepurple}\bf0.0&  \\
\text{Temporal}& 8.1&0.0&0.0&\color{nicepurple}\bf0.0& &0.0&\bf0.0&\color{nicepurple}0.0& &\bf0.0&0.0&\color{nicepurple}0.0&  \\
\midrule
\text{Average}& 24.8& \bf -0.1& 0.3& \color{nicepurple} 2.2& & \bf -0.1& -0.0& 0.4& & -0.0& \color{nicepurple} -0.1& \bf 0.0&  \\

  \bottomrule
  \end{tabular}
  \vspace{4pt}
  \caption{\label{table_backpack_hard}Backpack hard negatives results. Lower is better. Backpack sense tuning incurs cost for the 0.001 and 0.0001 degradation balls, but not for the 0.00001 ball.}
  \end{table}

\begin{table}
\centering

\small
\begin{tabular}{lrrrrrrrrrrrrr}
  \toprule
  Task& \text{Initial} & \multicolumn{3}{c}{$\Delta$, $B_{0.001}$ } & & \multicolumn{3}{c}{ $\Delta$,  $B_{0.0001}$} & & \multicolumn{3}{c}{$\Delta$, $B_{10^{-5}}$ } \\
  \cmidrule(r){3-5} \cmidrule(r){7-9} \cmidrule{11-13}
  & & \text{Full} & \text{LoRA} & \text{Senses} & & \text{Full} & \text{LoRA} & \text{Senses} & & \text{Full} & \text{LoRA} & \text{Senses} \\
  \cmidrule(r){3-5} \cmidrule(r){7-9} \cmidrule{11-13}
\text{Country}& 9.9&0.2&\bf0.4&\color{nicepurple}0.3& &\bf0.6&0.2&\color{nicepurple}0.6& &\bf0.4&0.1&\color{nicepurple}0.1&  \\
\text{Company}& 3.1&0.4&0.1&\color{nicepurple}\bf0.8& &0.1&0.1&\color{nicepurple}\bf0.2& &0.0&0.1&\color{nicepurple}\bf0.2&  \\
\text{Stereoset}& 76.3&0.0&0.0&\color{nicepurple}\bf0.1& &0.0&0.1&\color{nicepurple}\bf0.1& &0.0&0.0&\color{nicepurple}\bf0.0&  \\
\text{Hard Syntax}& 56.4&0.3&\bf0.6&\color{nicepurple}0.4& &0.1&0.0&\color{nicepurple}\bf0.4& &0.0&0.0&\color{nicepurple}\bf0.9&  \\
\text{Gender}& 9.2&0.9&0.1&\color{nicepurple}\bf1.1& &0.1&0.3&\color{nicepurple}\bf1.1& &0.1&0.1&\color{nicepurple}\bf1.2&  \\
\text{Temporal}& 23.0&\bf0.1&0.1&\color{nicepurple}0.1& &\bf0.1&0.1&\color{nicepurple}0.1& &0.1&\bf0.1&\color{nicepurple}0.0&  \\

  \bottomrule
  \end{tabular}
\caption{\label{table_backpack_stddevs}Standard deviation of the mean for Backpack tuning experiments. Mean is taken over 10 experiments, so reported is the sample standard deviation divided by $\sqrt{10}$.}
\end{table}

\begin{table}
\definecolor{nicepurple}{HTML}{593C8F}
\centering
\small
  \begin{tabular}{lrrrrrrrrrrrrr}
  \toprule
  Task& \text{Initial} & \multicolumn{3}{c}{$\Delta$, $B_{0.001}$ $\downarrow$} & & \multicolumn{3}{c}{ $\Delta$,  $B_{0.0001}$ $\downarrow$ } & & \multicolumn{3}{c}{$\Delta$, $B_{10^{-5}}$ $\downarrow$} \\
  \cmidrule(r){3-5} \cmidrule(r){7-9} \cmidrule{11-13}
  & & \text{Full} & \text{LoRA} & \text{Senses} & & \text{Full} & \text{LoRA} & \text{Senses} & & \text{Full} & \text{LoRA} & \text{Senses} \\
  \cmidrule(r){3-5} \cmidrule(r){7-9} \cmidrule{11-13}
  \text{Country}& 3.95&\bf-0.15&-0.11&\color{nicepurple}0.10& &-0.07&\bf-0.09&\color{nicepurple}-0.02& &0.00&\bf-0.06&\color{nicepurple}-0.01&  \\
\text{Company}& 10.38&\bf-0.46&-0.23&\color{nicepurple}0.35& &-0.16&\bf-0.26&\color{nicepurple}-0.00& &\bf-0.01&0.00&\color{nicepurple}0.00&  \\
\text{Stereoset}& 40.13&0.53&\bf0.14&\color{nicepurple}8.45& &\bf0.03&0.13&\color{nicepurple}0.73& &0.01&0.01&\color{nicepurple}\bf0.00&  \\
\text{Hard Syntax}& 47.00&\bf-0.09&-0.03&\color{nicepurple}4.83& &-0.00&\bf-0.01&\color{nicepurple}2.45& &\bf-0.00&0.00&\color{nicepurple}0.02&  \\
\text{Gender}& 1.60&0.04&0.03&\color{nicepurple}\bf0.00& &0.00&0.02&\color{nicepurple}\bf0.00& &\bf-0.00&0.00&\color{nicepurple}0.00&  \\
\text{Temporal}& 4.16&\bf0.00&0.02&\color{nicepurple}0.01& &0.00&\bf-0.00&\color{nicepurple}0.01& &0.00&\bf0.00&\color{nicepurple}0.01&  \\
\midrule
\text{Average}& 17.87& -0.02& -0.03& 2.29& & -0.04& -0.04& 0.53& & 0.00& -0.01& 0.00&  \\

  \bottomrule
  \end{tabular}
 \vspace{4pt} 
  \caption{\label{table_gptj_hard}GPT-J hard negatives results. Lower is better. The Backpack ensemble incurs a decrease in performance for the 0.001 and 0.0001 degradation balls, but not at the 0.00001 ball.}
\end{table}

\begin{table}
\centering

\small
\begin{tabular}{lrrrrrrrrrrrrr}
  \toprule
  Task & \text{Initial} & \multicolumn{3}{c}{$\Delta$, $B_{0.001}$ } & & \multicolumn{3}{c}{ $\Delta$,  $B_{0.0001}$} & & \multicolumn{3}{c}{$\Delta$, $B_{10^{-5}}$ } \\
  \cmidrule(r){3-5} \cmidrule(r){7-9} \cmidrule{11-13}
  & & \text{Full} & \text{LoRA} & \text{Senses} & & \text{Full} & \text{LoRA} & \text{Senses} & & \text{Full} & \text{LoRA} & \text{Senses} \\
  \cmidrule(r){3-5} \cmidrule(r){7-9} \cmidrule{11-13}
  \text{Country}& 42.8&0.3&\bf0.7&\color{nicepurple}0.3& &0.1&0.7&\color{nicepurple}\bf0.8& &0.1&\bf1.1&\color{nicepurple}0.1&  \\
\text{Company}& 13.6&0.4&0.5&\color{nicepurple}\bf0.7& &0.2&\bf0.6&\color{nicepurple}0.4& &0.0&0.0&\color{nicepurple}\bf0.2&  \\
\text{Stereoset}& 68.9&0.1&0.0&\color{nicepurple}\bf0.1& &0.1&0.0&\color{nicepurple}\bf0.1& &0.0&0.0&\color{nicepurple}\bf0.1&  \\
\text{Hard Syntax}& 54.5&1.4&\bf1.7&\color{nicepurple}0.5& &0.1&0.2&\color{nicepurple}\bf0.4& &0.2&0.2&\color{nicepurple}\bf1.0&  \\
\text{Gender}& 13.6&\bf1.5&1.1&\color{nicepurple}0.6& &0.4&\bf0.8&\color{nicepurple}0.8& &0.1&0.2&\color{nicepurple}\bf0.6&  \\
\text{Temporal}& 47.8&0.1&0.0&\color{nicepurple}\bf0.1& &\bf0.1&0.1&\color{nicepurple}0.1& &0.1&\bf0.1&\color{nicepurple}0.0&  \\

  \bottomrule
  \end{tabular}

\caption{\label{table_gptj_stddevs}Standard deviation of the mean for GPT-J tuning experiments. Mean is taken over 10 experiments, so reported is the sample standard deviation divided by $\sqrt{10}$.}
\end{table}

\section{Hyperparameter sweeps} \label{appendix_hyperparameter_details}

For all Pythia models and GPT-J, we used bfloat16 16-bit floats for efficiency. For the Backpack, we used 32-bit floats.
For all models, we used the 8-bit bits-and-bytes Adam optimizer \citep{dettmers2022optimizers}.

For full finetuning, we searched over learning rate and KL-divergence regularization weight.
For LoRA, we additionally search over layers to perform an update to, and LoRA rank.
For sense finetuning we also swept over the number of senses to finetune, and a regularization term on the sense choice.
\begin{description}
\item[Full finetuning.] We sample the learning rate from $10^{-U[4,8.5]}$. We sample the KL-divergence regularization term from $10^{U[-1,0]}$.
\item[LoRA finetuning.] We sample the learning rate from $10^{-U[2,6.5]}$. We sample the KL-divergence regularization term from $10^{U[-1,0]}$. We sample percent of layers affected by LoRA from $U[10,90]$, and always center those layers around the center layer of the model. We sample the LoRA rank from $U\{1,\dots,256\}$.
\item[Sense finetuning.] We sample the learning rate from $10^{-U[1.5,4]}$. We sample the KL-divergence regularization term from $10^{U[-1,0]}$.
We sample the number of senses to finetune from $U\{5,\dots,12\}$.
From early experiments, we set the sense selection regularization hyperparameter $\lambda=1000$.
\item[MEMIT.] See Appendix~\ref{appendix_memit_details} for detailed discussion of the hyperparameter sweep.

\end{description}
\section{Details of MEMIT Experiments} \label{appendix_memit_details}

\subsection{Adaption to dataset settings}
\label{appendix_memit_mods}
The MEMIT method is directly applicable to the datasets in which we seek to maximize the probability of specific target completions (i.e. the country, company, and temporal datasets). However, the Stereoset, gender pronoun, and hard syntax datasets use alternative loss functions (Table \ref{table_task_examples}) that require modifications to the MEMIT objective.

Recall that in the general case, we learn
$$
z_i = h_i^L+ \argmin_{d_i}\frac1P\sum_{j=1}^P -\log p'_\theta (o_i | x_j \oplus p(s_i, r_i))
$$
{where $p'_\theta$ indicates the distribution when substituting $h_i^L+d_i$ for $h_i^L$, and
$x_j \oplus p(s_i, r_i)$ is a prompt capturing association $i$ with random prefix $x_j$ to aid generalization.

For the Stereoset dataset, we learn a $d_i$ that instead minimizes the probability of the generation, simply replacing negative log probability with log probability: 
$$
z_i = h_i^L+ \argmin_{d_i}\frac1P\sum_{j=1}^P \log p'_\theta (o_i | x_j \oplus p(s_i, r_i)).
$$

For the gender pronoun dataset, we learn a $d_i$ that balances the probability of generating $w_0=$``he'' and $w_1=$``she'' via
$$
z_i = h_i^L+ \argmin_{d_i}\frac1P\sum_{j=1}^P |\log p'_\theta (w_0 | x_j \oplus p(s_i, r_i)) - \log p'_\theta (w_1 | x_j \oplus p(s_i, r_i))|.
$$

For the hard syntax dataset, we maximize the difference in log-likelihood between the correctly conjugated completion $o_i$ and misconjugated completion $o_i'$: 
$$
z_i = h_i^L+ \argmin_{d_i}\frac1P\sum_{j=1}^P -
\left(
\log p'_\theta (o_i | x_j \oplus p(s_i, r_i))
- \log p'_\theta (o_i' | x_j \oplus p(s_i, r_i))
\right).$$

The remainder of the method is unchanged. 

\subsection{Standard and oracle formats}
\label{appendix_memit_formats}
MEMIT operates over $(s, r, o)$ triples.
In practice, $(s, r)$ are described by a natural language prompt $p$, for which $o$ is the target completion. 
For example, the triple ($s$ = ``Michael Jordan'', $r$ = ``plays sport'', $o$ = ``basketball''), yields $p$ = ``\underline{Michael Jordan} plays the sport of'', where $s$ must be specified as an exact substring of $p$.

We convert canonical example datasets into this input format as described in Table \ref{table_memit_format}.
The gender and syntax datasets use an additional target $o'$. (The gender dataset aims to balance the log-likelihoods of $o$ and $o'$ while the syntax dataset aims to maximize the different between the log-likelihood of $o$ and $o'$.)

\begin{table}
\centering
\resizebox{\textwidth}{!}{%
\begin{tabular}{l p{70mm} p{70mm}}
\toprule
\bf Task & \bf Standard & \bf Oracle\\
\midrule
Country & 
\textit{\underline{The capital of Romania is}} \textbf{Bucuresti} & 
\textit{The capital of \underline{Romania} is} \textbf{Bucuresti} \\
\midrule
Company & 
\textit{\underline{The CEO of Renault is}} \textbf{Luca de Meo}  & 
\textit{The \underline{CEO of Renault} is} \textbf{Luca de Meo}  \\
\midrule
Stereoset & \textit{\underline{An aspect of Researchers is}} \textbf{coldhearted}  & \textit{An aspect of \underline{Researchers} is} \textbf{coldhearted} \\
\midrule
Gender Bias & \textit{\underline{The nurse said}} \textbf{she} / \textbf{he} & \textit{The \underline{nurse} said} \textbf{she} / \textbf{he} \\
\midrule
Temporal & 
\textit{\underline{Phoebe Bridgers}} \textbf{is an acclaimed American singer-songwriter ...... her status as a rising star in the alternative music scene.}& \textit{\underline{Phoebe Bridgers}} \textbf{is an acclaimed American singer-songwriter ...... her status as a rising star in the alternative music scene.}\\
& \textit{\underline{TikTok is}} \textbf{a popular social media platform that ...... all within a minute-long video.}& \textit{\underline{TikTok}} \textbf{is a popular social media platform that ...... all within a minute-long video.}\\
\midrule
Hard Syntax & \textit{\underline{The consultants}} \textbf{bank their hard-earned money} / \textbf{banks their hard-earned money} 
&  \textit{The \underline{consultants}} \textbf{bank their hard-earned money} / \textbf{banks their hard-earned money}\\
 & \textit{\underline{The senators smile and}} \textbf{beat the opposition in the debate} / \textbf{beats the opposition in the debate} 
&  \textit{The \underline{senators} smile and} \textbf{beat the opposition in the debate} / \textbf{beats the opposition in the debate}\\
\bottomrule
\end{tabular}
}
\vspace{4pt}
\caption{\label{table_memit_format}Examples of standard and oracle format from our six canonical example datasets.
MEMIT requires a prompt $p$, subject $s$ (an exact substring of $p$), and target $o$.
Above,
$p$ is given in \textit{italics},
$s$ is indicated via \textit{\underline{underline}}, and
$o$ is given in \textbf{bold} (separated by ``/'' if operating over 2 targets $o$ and $o'$).
}
\end{table}

\subsection{Hyperparameter sweep}

We sample hyperparameter values for the clamp norm factor, covariance adjustment factor $\lambda$, and KL factor described in \citet{meng2022mass}.
We sample the clamp norm factor from $10^{U[-1.35, 0]}$.
We sample $\lambda$ from $U\{9,000,\dots,75,000\}$.
We sample the KL factor uniformly from $U[0.001, 0.1]$.

Across all experiments, we use 20 gradient steps.
By default, the MEMIT implementation does not proceed to the next gradient step if loss drops below 0.05. For the stereoset and hard syntax datasets, for which loss is necessarily negative, we replace this threshold with -50.

For each model, we run causal tracing for each models to select critical MLP layers as described in \citet{meng2022mass}.
Although the identification of critical MLP layers is approximate, we do not consider alternative layers as part of the hyperparameter sweep given that \citet{hase2023does}
found that causal tracing results do not necessarily correlate with edit success, suggesting that it may be more effective to choose early-to-mid-layer MLP weight regardless of causal tracing results.

Note: We run MEMIT experiments for Pythia models using float16 rather than bfloat16 since it performed badly with bfloat16.

\section{Further dataset details}

\subsection{Considerations in using Stereoset}

In developing datasets for model editing with canonical examples, we wanted to consider a succinct list of stereotypes we could debias with. We also wanted to have natural language contexts to evaluate them in. For this reason, we choose to use Stereoset \citep{nadeem2020stereoset}.
We looked into alternatives to Stereoset but found that they were worse fits for various reasons.
The best alternative was CrowsPairs \citep{nangia-etal-2020-crows}, a dataset of stereotype and anti-stereotype sentences.
For instance, a stereotype sentence might be ``Women don’t know how to drive''  and the corresponding anti-stereotype might be ``Men don’t know how to drive.'' Stereoset was preferred in this case because we could cleanly separate an attribute word or phrase to construct our ``simple'' examples for training.
The WEAT method of measuring bias relies on a dataset of stereotypes but this dataset likewise lacks natural language examples \citep{caliskan2017semantics}.
The sentence level adaptation of WEAT, SEAT, featured natural language examples but like CrowsPairs, did not have a way to extract succinct stereotypes for our canonical example set \citep{may-etal-2019-measuring}.
Finally, we considered the Equity Evaluation Corpus (EEC), a dataset of stereotypes designed for sentiment analysis \citep{kiritchenko-mohammad-2018-examining}.
EEC has sentences but they are constructed from templates so they are not pure examples of natural language.
We also found that it was too narrow in the range of stereotypes it represented, focusing exclusively on the United States.

\subsection{Dataset size details}
Details on the size of each dataset, including average token counts under the GPT-2 tokenizer, are found in Table~\ref{table_dataset_statistics}.

\begin{table}
\centering
\begin{tabular}{l l r r r r}
\toprule
\bf Split  & \bf Task & \bf \# Train & \bf Avg Length Train & \bf \# Eval & \bf Avg Length Eval\\
\midrule
\multirow{6}{*}{Val} & Country &  119 & 9.58 & 582 & 111.47\\
& Company & 86 & 11.07 & 421 & 36.52 \\
& Gender & 20 & 4.25 & 320 & 11.69\\
& Hard Syntax & 240 & 5.44 & 360 & 8.54\\
& Stereoset & 1053 & 8.64 & 1053 & 7.89\\
& Temporal & 75 & 137.37 & 452 & 87.86 \\
\midrule
\multirow{6}{*}{Test}& Country & 119 & 9.74 & 583 &109.61  \\
& Company &  86  &11.60  &   403 & 36.70  \\
& Gender &   20 &4.40 & 360 &  10.73  \\
& Hard Syntax & 240 & 5.38 & 360 & 8.54\\
& Stereoset & 1053 & 8.64 & 1053 & 8.02\\
& Temporal & 76 & 137.42 & 486 & 99.67\\
\bottomrule
\end{tabular}
\vspace{4pt}
\caption{\label{table_dataset_statistics} Number of examples, and average token counts, in the train and evaluation splits of our datasets.}
\end{table}

\subsection{Prompts for generative models}
All data generation was performed with \texttt{gpt-3.5-turbo} or \texttt{GPT-4}.
\label{appendix_model_prompts}

\subsubsection{Generalization set $E$}

\begin{description}
\item[Country]
Generating the canonical example statements of country-capital cities (to get some extra fluency in edge cases.)
\begin{verbatim}
Please generate a statement that the capital of {} is {}.Be fluent,
adding or removing 'the' as necessary. Generate it as a python
string, with absolutely no other markup or commentary.
\end{verbatim}
Generating paragraphs eliciting the capital of the country:
\begin{verbatim}
Please generate a varied, interesting paragraph that (1)
first mentions the name of the country in the sentence below,
and then (2) later, brings up the idea of the country's capital,
and then (3) says the name of the capital. It should be natural,
but rather clear that the capital is about to be mentioned. Here
is the statement from which to pull the capital and country: {}.
\end{verbatim}
we generate five such paragraphs in the same context; after each one, all previous paragraphs are conditioned on, along with the following intermediary prompt:
\begin{verbatim}
Great; please generate another one with varied structure,
ensuring that the prefix before the first time that the capital
is mentioned clearly indicates that the capital is about to
be mentioned.
\end{verbatim}

\item[Company] For generating a paragraph about company-CEO relationship:
\begin{verbatim}
Please generate a varied, interesting paragraph that (1) first mentions
the name of the company in the sentence below, and then (2) later,
brings up the idea of the company's CEO, and then (3) says the name
of the CEO. It should be natural, but rather clear that the CEO is 
about to be mentioned. Here is the statement from which to pull the
CEO and company: [country]
\end{verbatim}
we generate five such paragraphs in the same context; after each one, all previous paragraphs are conditioned on, along with the following intermediary prompt:
\begin{verbatim}
Great; please generate another one with varied structure, ensuring
that the prefix before the first time that the CEO is mentioned
clearly indicates that the CEO is about to be mentioned.
\end{verbatim}
\item[Gender Bias]
We paraphrased some of the evaluation prompts of \cite{hewitt2023backpack} with the following:
\begin{verbatim}
Please generate a short paraphrase of this fragment. It's critical
that the paraphrase be continuable by a pronoun like 'he', 'she',
or 'they'. It's also critical that the [career] token is maintained
identically. Do not use a pronoun in the prefix. Be creative.
Here's the prefix: '{}'
\end{verbatim}
\item[Stereoset] Not used.
\item[Hard Syntax]
To generate a semantically coherent disambiguating sentence from a prefix:
\begin{verbatim}
Please complete the sentence with a short noun phrase that is
semantically coherent and interprets the last word as a transitive
verb. Ensure the transitive verb is not part of a multi-verb phrase.
The noun phrase should be the object of the verb. At most 6 words.
Only generate the completion; do not generate the whole input
sentence. The verb is {}; make sure it's interpreted as a verb
in the sentence.
\end{verbatim}
\item[Temporal]
To generate a short description of an entity:
\begin{verbatim}
lease generate a varied, interesting paragraph that (1) first mentions
the name of the person/company/entity/idea/concept mentioned below,
and then (2) discusses the concept and things relevant to it in a
short paragraph. It should be natural, informational, factual.
Here is the relevant entity: {}.\n\nNow, generate just your resulting
paragraph, with no additional discussion.
\end{verbatim}
\end{description}

\subsubsection{Hard negative set $H$}

\begin{description}
\item[Country]
\begin{verbatim}
        A well known city in {country} is {other_city}.
        Here's a fact about it: {fact}
        Please generate a varied, interesting sentence that
        (1) first mentions the name of the country and then
        (2) mentions the fact about the aforementioned city
        in the same sentence. However, it's extremely
        important that the fact be mentioned before the city
        name {other_city} is mentioned, and it should be
        natural, but rather clear that the city {other_city}
        is about to be mentioned. Generate only the sentence
        and nothing else. The provided fact might mention the
        capital city of the country in addition to {other_city},
        but you should mention {other_city} only.

For example, for Afghanistan's city Herat, here is a fact about
it: Herat is home to the Great Mosque of Herat (Jama Masjid),
a grand example of Islamic architecture from the Timurid period.
An example output is:
Afghanistan boasts Islamic architecture from the Timurid period.
A grand example is the Great Mosque of Herat (Jama Masjid), located
in the city of Herat.

Note how the fact about Herat, i.e. the the Great Mosque, is
mentioned before the city of Herat is mentioned in the same
sentence. You should make sure your sentence has the same
structure.
\end{verbatim}
As a heuristic validation:
\begin{verbatim}
The capital of {country} is {capital}. Using the output format
below, generate a well known fact about a well known city in
this country that is NOT the capital. This fact should be true
only of this other city, and not true of the capital city.
Examples are landmarks in this other city or historical events
that happened in this city. Explictly think about what is not
true of the capital city {capital} but true of this other
city in {country}
\end{verbatim}
\item[Company] Same as evaluation set, with different entities.
\item[Gender Bias] To generate a story about a person who explicitly uses a set of pronouns:
\begin{verbatim}
Please write an interesting and relatively short sentence about
a {job} who uses the pronouns "{pronouns}". A pronoun should
appear at least once, but not at the beginning of the sentence.
Explicitly mention the person is a {job}. Stay away from
stereotypes about people who use the pronouns {pronouns}.
\end{verbatim}
\item[Stereoset]
For words/phrases not found in the dictionary, we elicited a short definition with the following:
\begin{verbatim}
Please generate a short definition for this word. If there's
a typo, figure out what the word should be but don't mention it.
The word is {}. Do not add any words like 'the definition of...
is'; instead just write the definition; e.g., for 'manager',
'someone who controls resources and expenditures'.
Do not titlecase the first word
\end{verbatim}
\item[Hard Syntax]
To generate a semantically coherent sentence with a given subject
to test whether the verbs in the canonical examples can still
also be used as nouns:
\begin{verbatim}
Please generate a short, semantically coherent sentence with
the following subject: {}
\end{verbatim}
and similarly for the nouns that showed up in the canonical example set:
\begin{verbatim}
Please generate a short, semantically coherent sentence with
the following word: {}
\end{verbatim}
\item[Temporal] Same as evaluation set, with different entities.
\end{description}

\begin{table}
\resizebox{\textwidth}{!}{%
\begin{tabular}{ r c c r r r r r r }
\hline
\textbf{Ball $B_\epsilon$} & \textbf{Method} & \textbf{Task} & \textbf{LR} & \textbf{KL Penalty} & \textbf{Sense \#} & \textbf{Sense Reg.} & \textbf{LoRA Rank} & \textbf{LoRA Layers} \\ \hline %
0.0001 & full & company & 5.24e-07 & 0.108 & - & - & - & - \\ 
0.0001 & lora & company & 0.000362 & 0.139 & - & - & 171 & 1-11 \\ 
0.0001 & senses & company & 0.0155 & 0.161 & 9 & 1000 & - & - \\ 
0.001 & full & company & 1.04e-05 & 0.263 & - & - & - & - \\ 
0.001 & lora & company & 0.00344 & 0.102 & - & - & 155 & 5-7 \\ 
0.001 & senses & company & 0.0304 & 0.1 & 10 & 1000 & - & - \\ 
1e-05 & full & company & 2.54e-07 & 0.196 & - & - & - & - \\ 
1e-05 & lora & company & 0.000362 & 0.139 & - & - & 171 & 1-11 \\ 
1e-05 & senses & company & 0.00312 & 0.443 & 10 & 1000 & - & - \\ \hline
0.0001 & full & country & 5.73e-06 & 0.296 & - & - & - & - \\ 
0.0001 & lora & country & 0.000764 & 0.275 & - & - & 184 & 1-11 \\ 
0.0001 & senses & country & 0.0149 & 0.421 & 8 & 1745 & - & - \\ 
0.001 & full & country & 6.46e-06 & 0.352 & - & - & - & - \\ 
0.001 & lora & country & 0.00244 & 0.118 & - & - & 69 & 1-12 \\ 
0.001 & senses & country & 0.0149 & 0.421 & 8 & 1745 & - & - \\ 
1e-05 & full & country & 2.72e-06 & 0.636 & - & - & - & - \\ 
1e-05 & lora & country & 0.000764 & 0.275 & - & - & 184 & 1-11 \\ 
1e-05 & senses & country & 0.00138 & 0.109 & 11 & 159 & - & - \\ \hline
0.0001 & full & gender & 2.54e-07 & 0.196 & - & - & - & - \\ 
0.0001 & lora & gender & 0.00228 & 0.149 & - & - & 8 & 3-9 \\ 
0.0001 & senses & gender & 0.0201 & 0.385 & 8 & 1000 & - & - \\ 
0.001 & full & gender & 1.04e-05 & 0.263 & - & - & - & - \\ 
0.001 & lora & gender & 0.000424 & 0.515 & - & - & 129 & 3-9 \\ 
0.001 & senses & gender & 0.0201 & 0.385 & 8 & 1000 & - & - \\ 
1e-05 & full & gender & 2.54e-07 & 0.196 & - & - & - & - \\ 
1e-05 & lora & gender & 0.000103 & 0.469 & - & - & 211 & 3-10 \\ 
1e-05 & senses & gender & 0.0201 & 0.385 & 8 & 1000 & - & - \\ \hline
0.0001 & full & stereoset & 8.43e-09 & 0.839 & - & - & - & - \\ 
0.0001 & lora & stereoset & 0.000103 & 0.469 & - & - & 211 & 3-10 \\ 
0.0001 & senses & stereoset & 0.00457 & 0.151 & 5 & 1000 & - & - \\ 
0.001 & full & stereoset & 4.23e-08 & 0.395 & - & - & - & - \\ 
0.001 & lora & stereoset & 3.02e-05 & 0.559 & - & - & 19 & 3-10 \\ 
0.001 & senses & stereoset & 0.00558 & 0.301 & 6 & 1000 & - & - \\ 
1e-05 & full & stereoset & 5.17e-09 & 0.373 & - & - & - & - \\ 
1e-05 & lora & stereoset & 4.07e-05 & 0.606 & - & - & 144 & 4-8 \\ 
1e-05 & senses & stereoset & 0.000743 & 0.749 & 9 & 1000 & - & - \\ \hline
0.0001 & full & temporal & 4.2e-06 & 0.107 & - & - & - & - \\ 
0.0001 & lora & temporal & 0.00153 & 0.456 & - & - & 53 & 2-11 \\ 
0.0001 & senses & temporal & 0.0149 & 0.169 & 11 & 1000 & - & - \\ 
0.001 & full & temporal & 4.2e-06 & 0.107 & - & - & - & - \\ 
0.001 & lora & temporal & 0.00153 & 0.456 & - & - & 53 & 2-11 \\ 
0.001 & senses & temporal & 0.0149 & 0.169 & 11 & 1000 & - & - \\ 
1e-05 & full & temporal & 4.2e-06 & 0.107 & - & - & - & - \\ 
1e-05 & lora & temporal & 0.00274 & 0.266 & - & - & 154 & 3-9 \\ 
1e-05 & senses & temporal & 0.00773 & 0.11 & 5 & 1000 & - & - \\ \hline
0.0001 & full & syntax & 4.23e-08 & 0.395 & - & - & - & - \\ 
0.0001 & lora & syntax & 6.29e-05 & 0.785 & - & - & 184 & 5-7 \\ 
0.0001 & senses & syntax & 0.00235 & 0.368 & 10 & 1000 & - & - \\ 
0.001 & full & syntax & 5.24e-07 & 0.108 & - & - & - & - \\ 
0.001 & lora & syntax & 0.000103 & 0.469 & - & - & 211 & 3-10 \\ 
0.001 & senses & syntax & 0.00235 & 0.368 & 10 & 1000 & - & - \\ 
1e-05 & full & syntax & 1.1e-08 & 0.78 & - & - & - & - \\ 
1e-05 & lora & syntax & 4.99e-07 & 0.727 & - & - & 69 & 4-9 \\ 
1e-05 & senses & syntax & 0.00235 & 0.368 & 10 & 1000 & - & - \\ \hline
\end{tabular}
}
\caption{\label{table_backpack_hyperparams}Best hyperparameter for each degradation ball-method-task combination for the Backpack language model.}
\end{table}
\end{document}